\documentclass[preprint,12pt]{elsarticle}

\usepackage{amsmath,amssymb,amsfonts,mathtools}
\usepackage{algorithmic}
\usepackage{graphicx}
\usepackage{textcomp}
\usepackage[table]{xcolor}
\everymath{\displaystyle}
\newcommand{\argmax}{\arg\!\max}
\usepackage{subcaption}
\usepackage{float}
\usepackage{longtable}
\usepackage{booktabs}
\usepackage{import}

\usepackage{xparse}
\usepackage{etoolbox}

\newcolumntype{M}[1]{>{\centering\arraybackslash}m{#1}}
\newcolumntype{N}{@{}m{0pt}@{}}

\def\BibTeX{{\rm B\kern-.05em{\sc i\kern-.025em b}\kern-.08em
    T\kern-.1667em\lower.7ex\hbox{E}\kern-.125emX}}

\renewrobustcmd{\bfseries}{\fontseries{b}\selectfont}
\renewrobustcmd{\boldmath}{}
\newrobustcmd{\B}{\bfseries}

\begin{document}

\begin{frontmatter}

\title{Uncertainty Maximization in Partially Observable Domains: A Cognitive Perspective}

\author{Mirza Ramicic}

\address{Artificial Intelligence Center\\
Faculty of Electrical Engineering\\
Czech Technical University in Prague\\
12135, Prague, Czech Republic\\
ramicmir@fel.cvut.cz}

\author{Andrea Bonarini}

\address{Artificial Intelligence and Robotics Lab\\
Dipartimento di Elettronica, Informazione e Bioingegneria\\
Politecnico di Milano\\
20133, Milan, Italy\\
andrea.bonarini@polimi.it}

\begin{abstract}

Faced with an ever-increasing complexity of their domains of application, artificial learning agents are now able to scale up in their ability to process an overwhelming amount of data; However this comes at a cost of encoding and processing an increasing amount of redundant information.
This work exploits the properties of learning systems, applied in partially observable domains, defined to selectively focus on the specific type of information that is more likely to express the causal interaction among the transitioning states of the environment.
Experiments performed under a total of 32 different Atari game environments show that adaptive masking of the observation space based on the \textit{temporal difference displacement} criterion enabled a significant improvement in convergence of temporal difference algorithms applied to partially observable Markov processes under identical reproducible settings.
\end{abstract}


\end{frontmatter}

\section{Introduction}
\label{introduction}
\label{s:intro}
Recent rapid developments in \textit{reinforcement learning} (\textit{RL}) rely on the ability to perceive and process a great surge of information collected through the interaction with real or simulated environments. With the evolution of sophisticated artificial sensory apparatus began the collective quest to improve the predictability of surrounding world dynamics by increasing the shear amount of data collected from it.
The data greedy approach worked and consequently gave rise to significant breakthroughs and applicability of \textit{deep reinforcement learning} (\textit{DRL}). More complex architectures of neural network function approximators coupled with the increase of computational power allowed temporal-difference \textit{RL} algorithms to achieve super-human level control in problems that were designed for the complexity and scale of human cognition, such as Atari games~\cite{mnih2013playing}, complex board games such as Go~\cite{silver2017mastering,silver2018general,vinyals2019grandmaster,schrittwieser2020mastering}
, and modern strategy games like Starcraft II~\cite{vinyals2019alphastar}.

The aforementioned breakthrough approaches worked in part because both artificial and biological learning systems rely on the premise that their environment will provide them with enough informational entropy to improve predictability, while supporting their predominant function: adaptation. This Darwinian attribute of learning is evident in biologically-inspired \textit{machine learning} mechanisms such as \textit{RL} in the way that artificial agents \textit{adapt} to their environment by creating and updating a policy $\pi$ that would ultimately select actions according to the maximization of the expected reward in the long run~\cite{sutton2018reinforcement}. The adaptation of a \textit{RL} agent by learning can be seen as a process of reducing the inherent unpredictability or \textit{entropy} of the constantly changing environment: as the agent learns, it becomes better at predicting the environment dynamics i.e. how the environment reacts to the actions performed upon it. The learned predictive power allows an agent to gradually select better actions, i.e., those that would yield higher returns or rewards~\cite{sutton2018reinforcement}. In this adaptive view of the learning process an artificial agent is reducing its "surprise", or \textit{entropy}, about its perception of the environment according to the \textit{free energy principle}~\cite{friston2010free}. Artificial \textit{RL} systems faced with zero entropy \text{state space} and zero entropy \textit{reinforcement function} would make learning useless: no potential uncertainty to reduce means that the system cannot learn.

Collecting more data from the environment by \textit{DRL} approaches means that the learning agent's state space encompasses more of the external world unpredictability providing the learning algorithms with more entropy ``fuel'' for learning. However, in most real world cases the amount of data collected from the environment is not linearly proportional to the overall entropy it yields: increasing the amount of perceived data also increases the chance of encoding highly predictable and redundant data in the agent's state representations. Since that same data needs to be fed back into the learning system through a limited bandwidth \textit{channel} we can look at presented challenges as a \textit{communication problem}.Thus, under a \textit{communication channel} assumption in RL an artificial learning agent forms a limited capacity communication channel between its perception (sensory input) and machine learning algorithm itself. The \textit{communication channel} assumption intrinsically brings an important notion of its optimization; This is exactly the problem that Claude Shannon and John Tukey tried to solve during their period at the Bell Labs which inevitably led to the cornerstone of the information theory: the famous work by Shannon~\cite{shannon1948mathematical}.
The communication problem the two engineers faced in a nutshell is getting as much of information (Shannon's bits) through a channel of limited capacity measured by Tukey's bits. Theoretically the ideal communication case would be if the transferred information Shannon bits were equal to Tukeys: We would have used the full potential of the channels bandwidth.

In ML approaches, however, the optimization of the \textit{communication channel} equates to using the channel spanning peception and learning algorithm in such a way that would be beneficial for the entire learning process.

The approach presented in this work addresses the issue of optimization of a limited bandwidth \textit{communication channel} between the agent's perception and its learning algorithm, asserting the importance of looking at the learning problem (artificial and biological) as essentially \textit{uncertainty greedy}. 
This proposal is based on exploiting this inherent, natural, informational dependence, which represents a characteristic of all learning processes. Instead of increasing the channel's \textit{bandwidth} in our quest to better describe the environment (so increasing the state-space dimension) the  goal of the proposed approach is to utilize the available \textit{channel} in a way that would maximize its ability to efficiently transfer the uncertainty or entropy of the perceived environment. The proposal relies on the simple, yet effective, concept of \textit{temporal difference displacement}({TDD}) criterion for state space masking, able to perform a selective filtering of the sensed state portions based on the amount of transitional information it carries. TDD approach borrows motion detection techniques, usually used in eliminating temporal redundancy during video compression. Before using the transition $(s,a,r,s')$ for the learning, based on the two transitional states ($s$ and $s'$) TDD produces a binary matrix able to mask-out the non-transitional information contained in them. The TDD reduced states are then integrated into their specific transition forming an optimized reusable learning block experience carrying in-itself only temporally correlated information.


\textit{TDDM}-based masking makes possible for agent's learning experiences to include the information needed to represent \textit{distinctions} among world states (i.e. transitional information), while eliminating the constant information, alleviating the overload of the learning algorithm approximation process. 

The proposed selective, attentive focus thus inevitably creates partially observable spaces from the perspective of learning agents;
It does so in such a way as so to improve their ability to discriminate among the world states based on their temporal relationships; this, in turn, supports for development of better policies by inducing much needed determinism into the system.

The experimental results reported in Section~\ref{sec:experiments} show that the \textit{active state space masking} according to the TDD assumption can significantly improve the convergence of the \textit{TD} learning algorithms defined over a \textit{partially observable Markov process} in a variety of complex and sensory demanding environments such as Atari games. However, the \textit{TDD} as a broader concept can be applied to variety of other ML approaches relying on hidden-Markov processes for world modeling.
The article is structured in an incremental way, with the two first sections providing the general context of looking at the problem in a \textit{specific way}, therefore building up a foundation for the approach.

\section{The big picture: getting the right context}
\subsection{The perception problem: finding the right sources of uncertainty}
\label{perception}
Looking at the nature of things through the lens of the \textit{free energy principle}~\cite{friston2010free} imposes a duality: on one side we have a tendency of the universe, i.e., our environment, to achieve the state of least energy expenditure, which is a high entropy one (in both informational and thermodynamic way), and, on the other side, learning adaptive systems, both biological and artificial, that fight against to this natural tendency to disorder. This fundamental disposition for \textit{learning for adaptation} was observed from low complexity biological forms such as worms~\cite{rankin2004invertebrate} and even organisms with no nervous systems~\cite{boisseau2016habituation}. Evolving from the simpler forms, the majority of biological systems have ever since improved their sensory apparatus and started maximizing their potential by the development of mechanisms that enable them to better cope with the abundance of surrounding entropy. The solution was simple: focus on a finite subset of the environment and further evolve techniques to \textit{process} data observed from it to exploit the full potential of the specific \textit{perception}. 

For example, in the \textit{animalia} kingdom the organisms have evolved a strong preference for detecting \textit{electromagnetic waves} as they proved beneficial in reducing the uncertainty about their immediate environment, which, in turn, provided them with the possibility of better adaptation. Focusing on a specific range of electromagnetic spectrum allowed the formation of a structure that that we now refer to as ``eye''. Over time, the biological systems evolved many types of sensory apparatus, but none of them conveyed as much entropy as visual information: most of the physical reality does not necessarily make disturbances in the air we could detect, or emit chemical compounds, but reflect electromagnetic waves and, more importantly, in a variety of different ways. Sounds and smells just do not give the possibility to differentiate the properties of the environment to provide a high entropy sensory input, visuals do. This surge of entropy acquired by the newly founded ability to extract information from the visible light spectrum made a huge evolutionary leap in the Upper Paleolithic era~\cite{csikszentmihalyi1992imagining}: the search to expand the domain of perception quickly became a search to improve its processing. Perception moved from a simple reactive collection of neurons existing even before early Cambrian era~\cite{nilsson1996eye} to the highly complex processing of visual information that now happens in a human brain. Certainly, the human sensory apparatus also improved in the evolution, but the evolution of mechanisms that process the data it can produce had a major role in rising to the Upper Paleolithic evolutionary boom~\cite{csikszentmihalyi1992imagining}. The crucial ingredient was there, making sense of it was another issue.

\subsection{The Quest for Complete Control}

From a biological perspective~\cite{campbell1974evolutionary} the function of learning as a reduction of uncertainty is to ensure the survival of a specific organism, its \textit{fitness} to the environment. This enables it to choose the most adequate set of possible actions for any given situation, while effectively avoiding the risks of blindly trying out an action that could be fatal. The process of uncertainty reduction in a dynamic environment provides a constant goal-directed drive for evolutionary behaviour.

The early cybernetic work of \cite{ashby1961introduction,conant1970every} views the goal of survival in a dynamic environment strictly as a control problem, in which the control is exerted by compensating for dynamic perturbations that make the system deviate from its goal: maintaining or increasing its fitness.

If we consider survival as a primary evolutionary goal, mediated by the agent's ability to adapt to a constantly changing environment as a strict Ashby's type control problem, we could be bound to Ashby's Law of Requisite Variety (\cite{conant1970every}): For an organism to converge to an optimal evolutionary goal or complete control, in Ashby's view the variety of compensatory actions that its control system is capable to execute must be able to cope with the perturbations that might occur in its environment.

However, moving from a simplified control perspective, an adaptive learning system mediates this quest for complete control by creating and constantly updating a model of the objective environment. The infinite variety of all possible perturbations of the simple Ashby's type of system are now mapped onto a finite set of action-triggering representations: its state space.

An artificial learning agent moves from the one-to-one perturbation-action assumption of (\cite{conant1970every}) to a many-to-many type of a relationship between finite sets of state and action pairs. The evolutionary-driven fitness realization shifts from a simple reactive compensatory role to a much demanding long-term survival-supporting role of forming effective internal representations capable of effectively representing the features that are relevant to the agent's goal. The focus on agent perception ability within the machine learning community began mostly with the introduction of perceptual aliasing term by \cite{whitehead1990active,whitehead1991learning} which would be later incorporated in the wider concept of \textit{active perception}.

An actively perceiving artificial learning agent capable of effectively conveying the variety and importance of possible perturbations of the system through its internal representations while sustaining its discriminatory ability is capable of producing action polices that can exert an adequate amount of control for the agent to be able to achieve its goal of adaptation. This work, under the \textit{communication channel} assumption, explores how and to which extent artificial learning agents can maximize the potential of their internal representations to deliver the information that would be most supportive for the learning goals.

\subsection{Not all entropy is useful: a qualitative perspective on information}
\label{s:cc}
Life's quest for a reduction of uncertainty (as far as we know) didn't appear in high energy environments such as the gas giants of our Solar system, nor did it sustain in the low energy ones such as the Earth's Moon or Mars, for example.

This biological process, however, has some prerequisites in terms of the amount of dynamic perturbations of an environment: they need to be just enough to enable to predict the patterns of \textit{causal} relationships between the changing  states according to the \textit{integrated information principle}~\cite{tononi2016integrated} focusing on the dynamic environmental processes that have a causal influence on the system's goal.

The \textit{integrated information} $\Phi$ represents the information that is \textit{irreducible} to its non-interdependent subsets, which, in our case, are the representations of the environmental states. Instead, this type of information explains the \textit{relationships} between them, supporting the integration of a set of phenomenal distinctions into a \textit{unitary} experience~\cite{tononi2016integrated}.

The ability of a learning system to extract the information from its environment depends on the amount \textit{causal}, different perturbations found in that environment which are meaningful to its learning goal. This selective process can again be viewed as a direct optimization of the agents' \textit{communication channel} between its perceptual and learning system.

A constantly varying and high-entropy environment, as a prerequisite for an evolutionary drive, entails an organism capable of not only mapping static environment objects to their representations, but also dynamic perturbations of the environment (such as temporal correlations) to its own actions.

The preference for cognitively mapping the environment dynamics that exhibits temporal correlations, rather than its static equilibrium states has been predominantly observed in insects perceptual mechanisms ~\cite{rosner2013widespread,kirchner1989freely,srinivasan2020vision}. The focus on the dynamic perturbations of the environment by the way of motion flow detection in honeybees~\cite{srinivasan2020vision} and locusts~\cite{simmons2010escapes} is essential for mediating their flight steering maneuvers and escape jumps in response to a looming threats.

The well known \textit{Goldilocks}~\cite{rampino1994goldilocks} thermodynamics property of habitable planets can be extended in the Shannon's sense as the \textit{optimal informational saturation} condition of all learning systems, biological and artificial.

Thus, a more perceptually efficient agent could, under the \textit{communication channel} assumption,  mediate and alias the perception to a greater extent, while optimizing its communication channel, effectively allowing the learning system itself to focus on a more "useful" information (i.e. the information carrying more of the environments' dynamic perturbations~\cite{tononi2016integrated}).

Under cognitive perspective the efficacy of learning not only depends on the level of environment entropy, but also on the amount of the entropy that can be perceived or channeled to the learning system itself. This property provides a basis and has been heavily exploited under the \textit{communication channel} assumption introduced in Section~\ref{s:intro}


\subsection{Learning to live with uncertainty (and learn from it)}

The breakthroughs in artificial learning algorithms mentioned in Section~\ref{s:intro} have dealt with the \textit{uncertainty} of the world by focusing on the part of it that was \textit{deterministic} in nature and defining it as a \textit{Markov Decision Process} (\textit{MDP})~\cite{sutton2018reinforcement}. This represented a sort of a \textit{leap of faith} as most of real-world problems are inherently non-Markovian: the world itself is highly non-Markovian, and complex biological learning systems like humans have benefited from this as suggested in Section~\ref{perception}.
Even though, since the mid 60ies the artificial intelligence community have developed methods that could represent and reason with uncertainty that originate from the \textit{control engineering} perspective of Karl Johan Åström~\cite{aastrom1965optimal}. The majority of \textit{TD} methods have relied on this deterministic \textit{safe haven} of \textit{MPDs}. This tendency could be partially attributed to the fact that the proofs of the convergence of \textit{TD} algorithms assumed the agent's perceived state space to be Markovian and ergodic in nature~\cite{watkins1992q,tsitsiklis1994asynchronous}.
Despite the convergence issues the \textit{artificial intelligence} community adopted a non-deterministic method
as a (more or less) natural extension of \textit{MDP} under the name of \textit{partially observable Markov decision process} or POMDP~\cite{monahan1982state,lovejoy1991survey,cassandra1994acting}.

\subsection{Extending the MDP}
A \textit{Markov Decision Process} is fully defined by the tuple $\langle S,A,T,R \rangle$, which includes: a finite set of environment representations \textit{S} that can be \textit{reliably} encoded by the agent, a finite number of actions \textit{A} that an agent is allowed to perform in that environment, a transitional model of the environment \textit{T} providing a functional mapping of $S \times A$ to discrete probabilities defined over $S$, and a reward function $R(s,a)$ which maps the state and action pairs from $S$ and $A$ to a scalar indicating the immediate reward feedback the agent receives from being in a specific state $s$ and taking a specific action $a$~\cite{sutton2018reinforcement}.
In \textit{POMDP's} the algorithm doesn't have the benefit of performing the mappings of $S \times A$ over a set of deterministic states $S$ but rather on a set of the possible partial observations \textit{O} of the states~\cite{cassandra1994acting}. In other words, the additional modelling of the concept of \textit{observation} was required. The solution for this problem came in the form of a \textit{belief state}: an internal representation that maps the environment states to the probability that the environment \textit{is} actually in that state. The \textit{belief state} denoted by $B$ is simply a probability distribution that can be represented by a vector of probabilities, one for each possible state of the environment, summing to 1~\cite{cassandra1994acting}. This articulates the problem of learning in a partially observable environment as a problem of \textit{estimating} the "true" state of the world based on the \textit{belief state} derived from the agent's \textit{partial} observations. 

The \textit{POMDP} agent improves its estimates of the model of the environment by updating its \textit{state estimate} $\tau(b,a,o,s')$ 
about the state $s'$ based on the previous belief state $s$ along with the most recent action $a$ and the most recent \textit{partial} observation $o$ by applying the simplicity of the Bayes' rule according to Equation~\ref{eq:belief}. Transitional probabilities $\tau(s,a,s')$ in Equation~\ref{eq:belief} are given as in vanilla \textit{MDP's} and $b(s)$ represents the actual probability that is assigned to the state $s$ considering an agent being in a specific belief state $b$.

\begin{equation}
\begin{split}
\tau(b,a,o,s') & = P(s'\vert a,o,b) \\
                   & = \frac{P(o\vert s',a,b)P(s'\vert a,b)}{P(o\vert a,b)} \\
                   & = \frac{O(s',a,o)\sum_{s\in S}\tau(s,a,s')b(s)}{P(o\vert a,b)}
\end{split}
\label{eq:belief}
\end{equation}

Regardless of their differences, solving problems defined over \textit{MDP} and \textit{POMDP} come down to finding a policy $\pi$ that will maximize the future expected reward~\cite{sutton2018reinforcement}. While in the case of \textit{MDP} this policy represents a mapping of deterministic states $S$ to actions, in \textit{POMDP} the actions are chosen based on the basis of the agent's current \textit{belief states} $b$. Along the iterative update of the agent's \textit{belief state} using Equation~\ref{eq:belief} the agent's first step towards learning a policy $\pi$ is the iterative update of the \textit{value} functions $V$ for each of its belief states using \textit{dynamic programming} methods~\cite{bellman1966dynamic} such as \textit{value iteration} outlined in Equation~\ref{eq:bellman} (this is where the inherent complexity of the POMDP approach becomes apparent). The updated \textit{value} function $V_{n+1}$ in Equation~\ref{eq:bellman} is calculated on the basis of the previous \textit{value} function $V_n$ defined over the current \textit{state estimate} given by Equation~\ref{eq:belief} and immediate expected reward $r(b,a)$ of executing action $a$ in belief $b$. The expectation of this scalar reward $r(b,a)$ is based on the whole \textit{state space} and the current belief $b(s)$ as defined in Equation~\ref{eq:reward}.

\begin{equation}
V_{n+1}(b) = \max_{a}\left[ r(b,a)  + \gamma \sum_{o\in O}P(o\vert b,a)V_n(\tau(b,a,o))\right] \forall b \in B
\label{eq:bellman}
\end{equation}

\begin{equation}
r(b,a) = \sum_{s\in S}b(s)r(s,a)
\label{eq:reward}
\end{equation}

For an arbitrary \textit{value} function $V$ updated by Equation~\ref{eq:bellman}, a policy $\pi$ is said to be actually \textit{improving} on V under the Equation~\ref{eq:policy}
. The convergence of the policy $\pi$ to the optimal policy $\pi^*$ is the result of the value function $V$ convergence to $V^*$ as the number of iterations $n$ goes to infinity.

\begin{equation}
\pi(b) = \argmax_a \left[ r(b,a)  + \gamma \sum_{o\in O}P(o\vert b,a)V(\tau(b,a,o))\right] \forall b \in B
\label{eq:policy}
\end{equation}

If we simply omit the $\max_{a}$ operator from Equation~\ref{eq:bellman} we get a representation of the \textit{value} of executing a specific action $a$ in a current belief state $b(s)$. This representation, also known as Q-value, is given in Equation~\ref{eq:bellman-q} and its is widely used in temporal-difference learning since Watkins' paper~\cite{watkins1992q}.

\begin{equation}
Q_{n+1}(b,a) = \left[ r(b,a)  + \gamma \sum_{o\in O}P(o\vert b,a)V_n(\tau(b,a,o))\right] \forall b \in B
\label{eq:bellman-q}
\end{equation}

\subsection{POMDP as state-of-the-art}
However, solving a problem defined over \textit{POMDP} proved not to be such an easy task due to the complexity~\cite{ross2008online,lee2008makes} imposed by a creation of a \textit{belief state} $B$ that, in most of the cases, has the same dimension as $\vert S \vert$: the dimension of the belief space thus grows exponentially with $\vert S \vert$.
A certain revival for POMDP's, though, came with the introduction of more complex function approximators: for most non-trivial cases keeping track of the values $V$ given by Equation~\ref{eq:bellman} for each of the observations was computationally unfeasible because of their sheer numbers, and for this reason the value functions have been approximated by \textit{ANN} ranging back to Lin~\cite{lin1991programming}.
The approximation is done by nudging the parameters or weights $\Theta$ of an ANN by a small learning rate $\alpha$ at each learning step so that the current estimate $Q(b,a;\Theta)$ will be closer to the \textit{target} Q-value given by Equation~\ref{eq:bellman-q}. This is done by minimizing the loss function $L(\Theta)$ representing the difference between the previous estimate and the expectation \textit{target} by performing a \textit{stochastic gradient descent} on the weights $\Theta$ to achieve $ Q(s,a;\Theta) \approx Q^*(s,a) $ according to Equation~\ref{eq:gradient}:

\begin{equation} \label{eq:gradient} \nabla_{\Theta_i}L_i(\Theta_i) = 
\left ( y_i - Q(b,a;\Theta_i) \right )\nabla_{\Theta_i}Q(b,a;\Theta_i),
\end{equation}

where $y_i$ represents our target Q-value obtained by calculating the Bellman optimality under the newly observed transition parameters over Equation~\ref{eq:bellman-q}.

Later, the introduction of many-layered \textit{deep} ANN's capable of scaling up to an over-increasing sensory demand of modern RL applications~\cite{mnih2013playing,mnih2015human,silver2017mastering,silver2018general,vinyals2019alphastar,vinyals2019grandmaster,schrittwieser2020mastering} inspired \textit{deep learning} \textit{POMDP} approaches by Hausknecht and Stone~\cite{hausknecht2015deep} and, more recently, by Le~\cite{le2018deep}. Their \textit{deep recurrent Q-network} (\textit{DRQN}) achieved a better adaptation of agents under the circumstances where the quality of observation changes over time compared to vanilla \textit{DQN}.
Because of its appoximation power and scalability over the \textit{POMDP} domain the \textit{DRQN} approach is used as a basis for the proposed \textit{TDDM} filtering architecture further elaborated in Section~\ref{section:architecture}.

\section{POMP as Perception Mechanism}

Why would we link perception to \textit{partially observable Markov decision processes}? The nature of perception itself lies in selective filtering, processing, redefining, and, in most cases, interpreting the raw data received through an agent's sensory apparatus.

An agent is not acting upon the idealized full potential of the informational content present in its immediate environment, but on a small subset of processed observations, which are often moved to latent spaces. Despite of this limitation, the biological agents may act optimally in a partially observable world by building something that could be seen as POMDP belief states.

\subsection{Pioneering approaches}
One of the first \textit{computational} perspectives on perception was concerned about its most dominant and entropy rich modality: vision as formulated in the late 70ies by David Marr~\cite{marr1976early,marr1982vision}. These works postulate a theory of early visual computational processing that has inspired some of the pioneering works~\cite{agre1987pengi,agre1988dynamic} dealing with the problem of computational perception as an important component of artificial learning agents.
Marr's work set a theoretical base for the principle of what Agre and Chapman called \textit{deictic representation} by postulating that the first operation on a perceived raw image is to transform it into a more simple, but entropy rich, description of the way its intensities \textit{change} over the visual field, as opposed to a description of the intensities themselves~\cite{marr1976early,marr1982vision}. This \textit{primal sketch}, as he coined it, provides a description of significantly reduced size that is still able to preserve the important information required for image analysis.
The importance of the Agre and Chapman \textit{deictic} approach~\cite{agre1987pengi,agre1988dynamic} from the perspective of the here proposed work lies in its architecture: the \textit{crisp} distinction between the \textit{ perception} system and the \textit{central} system (i.e. the learning algorithm). The \textit{visual} system thus takes the \textit{deictic} burden: at any given moment, the agent's representation should actively register only the features or information that are relevant to the goal and \textit{ignore} the rest.
This architectural modularity allows the \textit{central} system in~\cite{agre1987pengi,agre1988dynamic} to be implemented in a rather simple way without the complexity of a pattern matcher or similiar computationaly demanding processes; the \textit{deictic} process enables generalization over functionally and indexically identical states of the environment by simply \textit{not bringing in} the \textit{redundant} distinctions among them.

Later work of Ballard et. al~\cite{ballard1997deictic,hayhoe1997task} put the \textit{deictic} principle of~\cite{agre1987pengi,agre1988dynamic} into the broad context of visual processing of biological systems suggesting that the human visual representations are \textit{limited} and \textit{task dependent}. \cite{ballard1997deictic,hayhoe1997task} further postulate that the superior human performance in visual perception can be attributed to the sequence of constraining \textit{deictic} processes based on a limited amount of primitive operations supporting the notion that a human working memory is limited in its capacity and computational processing ability~\cite{broadbent19581958,baddeley1992working,salway1995visuospatial}.

A more complex extension of~\cite{agre1987pengi,agre1988dynamic} is given by Chapman~\cite{chapman1992intermediate} through their \textit{SIVS} architecture. \textit{SIVS} have introduced a selective \textit{deictic} visual processing of subsets of an image by identifying the regions that are "task dependent". The interesting part of the \textit{SIVS} approach is that it implements, amongst other, a concept of \textit{visual routines} inspired by~\cite{ullman1987visual}, which actively process the visual information within the \textit{time-domain}, allowing for the detection and abstraction of \textit{changes} in the visual field~\cite{chapman1992intermediate}. The Chapman's applications of the \textit{visual routines} is very much in line with the temporal context retaining properties of \textit{POMDP}-based learning algorithms presented in this work.

\subsection{Getting the Problem Right}
The \textit{deictic} way of looking at a machine learning problem seemed very promising because of its ability to represent as \textit{equivalent} the world states that require the same action according to the agent's current policy: more abstracted, more compacted representations reduce the burden on learning mechanisms. As the researches eagerly exploit the possibilities of modeling artificial perception under the \textit{deictic} principle a concern arises whether this selective, compact, and task-dependent world representation can be acted upon \textit{deterministically} with respect to the \textit{Markov} property in order for an agent to achieve optimal policy~\cite{whitehead1991learning,chrisman1991intelligent}. The integration of adaptive control methods such as active perception with the machine learning algorithms~\cite{watkins1992q} may lead to a phenomenon of \textit{perception aliasing}~\cite{whitehead1991learning} as it can produce internal representations that are not \textit{consistent} with each other. 

Lack of \textit{consistency} among states can be very detrimental to the \textit{TD} algorithms~\cite{watkins1992q} as their underlying principle of Bellman's optimality~\cite{bellman1966dynamic,sutton2018reinforcement} relies on this property: inconsistent states can destabilize the learning algorithm by introducing unfounded maximums in the value function~\cite{whitehead1991learning} which, in turn, can make the agent diverge from its optimal policy. Furthermore, \textit{perceptual aliasing} can lead to \textit{distinct} world states that may call for equally \textit{distinct} actions according to the optimal policy being represented by the same \textit{deictic} representation.

\subsection{Correlation Saves the Day}
A \textit{partial} solution was readily proposed by the work that introduced the problem of \textit{perceptual aliasing}~\cite{whitehead1991learning} in the first place and it was based on detecting and suppressing the representations that are less correlated. As the \textit{MDP} assumption still relied on the deterministic principles the correlation between the states in~\cite{whitehead1991learning} seemed to be a part of the system that was the source of certainty. 

The here presented work extends that notion in the direction of the work by~\cite{chrisman1992reinforcement} that used the \textit{memory} of the previous states in order to detect the essential information that can induce correlations. In this case, the previously experienced correlations among the states are used to build a \textit{probabilistic model} that is able to \textit{predict} the current world state. Although probabilistic models have been used in reinforcement learning as a form of experience replay~\cite{sutton1991dyna,lin1991programming} the so-called \textit{predictive distinctions approach} of Chrisman~\cite{chrisman1992reinforcement} used it to drop the deterministic assumptions of the agent's representations by implementing a \textit{Hidden Markov Model}~(\textit{HMM})~\cite{rabiner1986introduction}. 
Proposing the powerful, yet (at the time) untapped predictive ability of \textit{RNN's}~\cite{jordan1992forward} to extend the Chrisman's approach led to artificial agents with a better grasp of uncertainty which, in turn, led to the definition of \textit{POMDP}.

McCallum~\cite{mccallum1993overcoming} used a similar \textit{HMM} approach, so-called \textit{utile distinction memory}, introducing the possibility of discriminating states based on their perceived \textit{utility}: the world states represented by the identical observations could be distinguished based on their prior assignments of rewards thus driving the system ability to discern the states. The \textit{utile distinction memory}~\cite{mccallum1993overcoming} approach raised a possibility for further optimizations of the memory process itself as seen in the later work by Wiestra and Weiring~\cite{wierstra2004utile} on \textit{utile distinction hidden Markov models} or \textit{UDHMM}. We can relate the \textit{UDHMM} approach to the here presented work as it too optimizes the learning process by limiting the amount of the informational entropy being channeled to the learning algorithm in such a way that distinctions of the specific world states would be represented in memory \textit{only} when needed.

While the \textit{UDHMM} does this by adjusting the number of steps it looks back in order to create the \textit{utility} distinctions, the approach proposed in this work rather focuses on isolating a \textit{subset} of the observations that induce the distinction-relevant information while ignoring the extraneous part. This observation partitioning principle has been successfully implemented in a class of \textit{POMDP's} called \textit{mixed observability Markov decision process} or \textit{MOMDP} introduced by Ong et al.~\cite{ong2009pomdps}. \textit{MOMDP} exploits the fact that although the agent perceives limited representations of the world, some subset of its observations can be deterministic in the sense that they possess a \textit{fully observable} property. In \textit{MOMDP} approach the agent state is split into the fully observable component $x$ and the partially observable one $y$ which leads to the computational benefits of maintaining and updating a \textit{belief state} $b_y$ about the $y$ component only.

The aforementioned improvements of the base algorithm bring the focus on the problem of \textit{artificial perception}~\cite{weyns2004towards,spaan2008cooperative} as a way for an agent to intrinsically and dynamically learn \textit{what} to perceive in the first place. One of popular approaches to active perception defined over \textit{POMDP} includes designing a \textit{reinforcement} function in such a way that it would minimize the sensing cost~\cite{boutilier2002pomdp}, minimize the agent's \textit{belief state} uncertainty based on its current measure of entropy~\cite{araya2010pomdp}, or credit the belief level achieved by the specific sensed state~\cite{spaan2015decision}.

To mitigate the effects of encoding a great amount of low-entropy data that does not support the learning process, the recent approaches prioritized on the agent's experiences that carried more entropy~\cite{ramicic2017entropy} or used an array of unsupervised learning techniques in order to compress the world representations into vectors with high entropy~\cite{vinyals2019alphastar}. Both approaches were effectively conveying more of the environment's uncertainty to the learning agent itself, improving the learning performance.

More recent work~\cite{zhu2017improving} relates observations with the agent's actions by encoding them together in such a way that the \textit{LSTM} layer can propagate the additional context of \textit{actions} through the history of \textit{observation-action} pairs.
In~\cite{de2019influence}, ANN function approximators are used to split the sensory input in the partially observable subset that is included in the \textit{POMDP's} history of the past states and the \textit{fully observable} subset that is treated as Markovian.

Artificial attention has been explored recently in the context of standard MDP-based reinforcement learning problems through evolutionary techniques, taking biological inspirations such as intentional blanking in the approach by Tang et. al.~\cite{tang2020neuroevolution}.

\section{Model Architecture and Theoretical Background}
\label{hypothesis}
This work introduces a novel method of improving the propagation of environment's inherent uncertainty or entropy to the \textit{temporal-difference} reinforcement learning algorithm defined over a \textit{partially observable domain} by introducing a perceptual \textit{aliasing} method of the agent's state space based on the concept of \textit{temporal difference displacement} criterion or \textit{TDD}. The \textit{TDD} perceptual aliasing acts by optimizing the \textit{communication channel} established between perception and the learning mechanism in such a way as to transfer as much of the environments dynamic perturbations, or causal information~\cite{tononi2016integrated} possible given the channels limited bandwidth.

The \textit{TDD} criterion as perceptual aliasing mechanism selects the information perceived by the agent, so that the information which does not contain learning potential (as defined by the \textit{communication channel} assumption presented in Section \ref{s:cc}) will not be transferred to the learning algorithm.

The proposed \textit{TDD} criterion exploits the following properties of these types of learning algorithms:
\begin{itemize}
    \item By each successive \textit{TD} \textit{transition} the algorithm takes advantage of the \textit{temporal relations} between the transitioning states in order to improve its \textit{belief state} about the environment: the \textit{policies} that an agent develops are not a product of  \textit{deterministic} states, but are based on the history of (possibly) all previous observations and their underlying relationships.
    \item The \textit{POMDP} agent still updates its \textit{policy} based on a \textit{single} transition from state $s$ to state $s'$ by performing an action $a$: this \textit{one step} information along with a reward scalar constitutes everything the algorithm needs in order to perform a learning update~\cite{watkins1992q}.
\end{itemize}

Moreover, the \textit{TDD} criterion \textit{postulates} the following:

\begin{itemize}
    \item Each of the two subsequent states ($s$ and $s'$) in a \textit{single} learning transition can either have a positive or negative effect on the uncertainty reduction of the \textit{belief state}, based on their temporal \textit{relationship}~\cite{cassandra1994acting}.
    \item The modification between subsequent \textit{observation states} ($s$ and $s'$) that a specific transition has induced is as relevant to reduce uncertainty about the agent's \textit{belief state} as its informational content supports the ability to \textit{distinguish} states from each other.
    \item The changes in the \textit{observation states} ($s$ and $s'$) channel a lot of the environment uncertainty or entropy required for creating an accurate \textit{observation model}: they carry the highly valuable information about the \textit{transitional relationships}.
    \item For the intuition regarding the previous paragraph let us imagine a case in which all of the observations were exactly the same but the transitions yield different rewards. The \textit{POMDP} learning algorithm would try to attribute the reward differences to the states in the form of \textit{value functions}, but there would not be learning because the states would be indistinguishable from each other ($s = s'$).
\end{itemize}

In other words \textit{TDD} provides a simple yet effective way to maximize the amount of informational entropy that is dedicated to the representation of causal relationship between the environment states or states representations according to the integrated information principle~\cite{tononi2016integrated}.

\begin{figure}[h!]
    \centering
    \includegraphics[width=0.66666\linewidth]{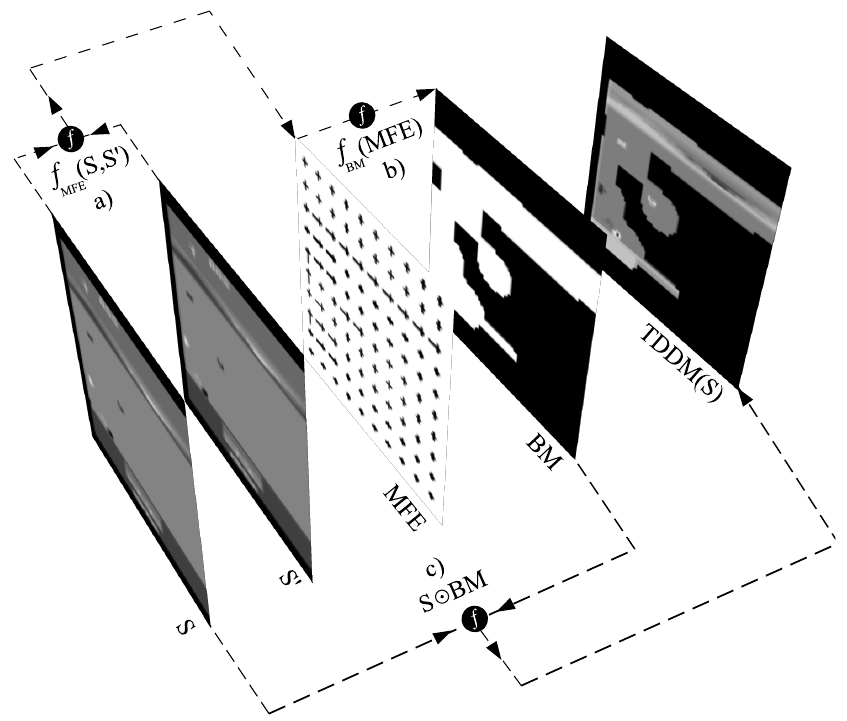}
    \caption{The process of active state masking based on \textit{TDD} decomposed with regards to its functional transformations $f$ represented by black circles; a) motion estimation based on polynomial expansion \cite{farneback2003two}; b) Binary threshold mask generated from the motion-field vector magnitudes obtained from a); c) Element-wise matrix multiplication of the original frame S with the binary mask obtained from b).}
    \label{fig:filter}
\end{figure}

The full potential of the \textit{TDD} perspective on learning is realized through an \textit{active state space masking} or selective filtering of the agent's observations based on the \textit{temporal difference displacement} between the initial perceived state $s$ and its successor $s'$. Figure~\ref{fig:filter} details the applied \textit{TDD} transformations to an Atari game learning problem example.

In the visual channel used in the examples on which we tested the systemd, the \textit{TDD} criterion is estimated with a computationally inexpensive two-frame motion estimation technique based on polynomial expansion~\cite{farneback2003two} capable of producing a dense optical flow vector field based on two successive video frames, which, in the case of \textit{TDD} includes observation states of the agent's atomic transition ($s$ and $s'$) as detailed in Figure~\ref{fig:filter}.

After the initial problem-specific prepossessing of perceived visual information, the two successive frames, namely, $S$ and $S'$ are used as an input for the motion field estimation function $f_{MFE}$ in Figure~\ref{fig:filter} $a)$.

In order to perform the motion estimation $f_{MFE}$ function analyses the displacement of the intensities (dx,dy) between the starting image $I(x,y,t)$ at the time $t$ and image $I(x+dx,y+dy,t+dt)$ that is obtained after temporal displacement $t+dt$.

The initial displacement analysis~\cite{farneback2003two} produces a dense motion field estimate window or MFE (see Figure~\ref{fig:filter}) commonly depicted by using oriented Cartesian vectors representing the intensity and direction of detected temporal displacements.

The obtained dense motion field is then transformed to a binary threshold mask or BM in Figure~\ref{fig:filter} by applying a simple adaptive high-pass filter on the magnitude component of the vector.

Each transition generates its own~\textit{unique} binary mask BM which is multiplied \textit{element-wise} with each state input in Figure~\ref{fig:filter}c) effectively performing TDD masking $TDDM(S)$ on the input prior to its integration into the main TD learning algorithm.

\begin{figure}[h!]
    \centering
    \includegraphics[width=0.66666\linewidth]{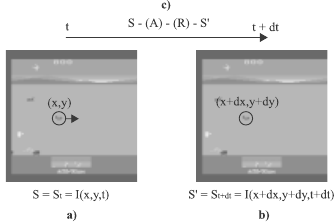}
    \caption{Simplified process of motion estimation based on the amount of displacement $(dx,dy)$ detected between two transitional states of an Atari game example. The common Atari preprocessing includes resizing the input to a 84x84 matrix and reducing three color channels to a single grayscale one;  a) Image intensity $I(x,y,t)$ at time $t$ or $S_t$; b) Image intensity $I(x+dx,y+dy,t+dt)$ after a $dt$ amount of time has passed or $S_{t+dt}$; c) During the $dt$ time-window the autonomous agent has successfully performed a transition defined over a MDP by taking an action $(A)$, obtaining immediate reward $(R)$ and observing $S_{t+dt}$ state at the final time of $t+dt$.}
    \label{fig:transiton}
\end{figure}

\begin{figure}[h!]
    \centering
    \includegraphics[width=0.66666\linewidth]{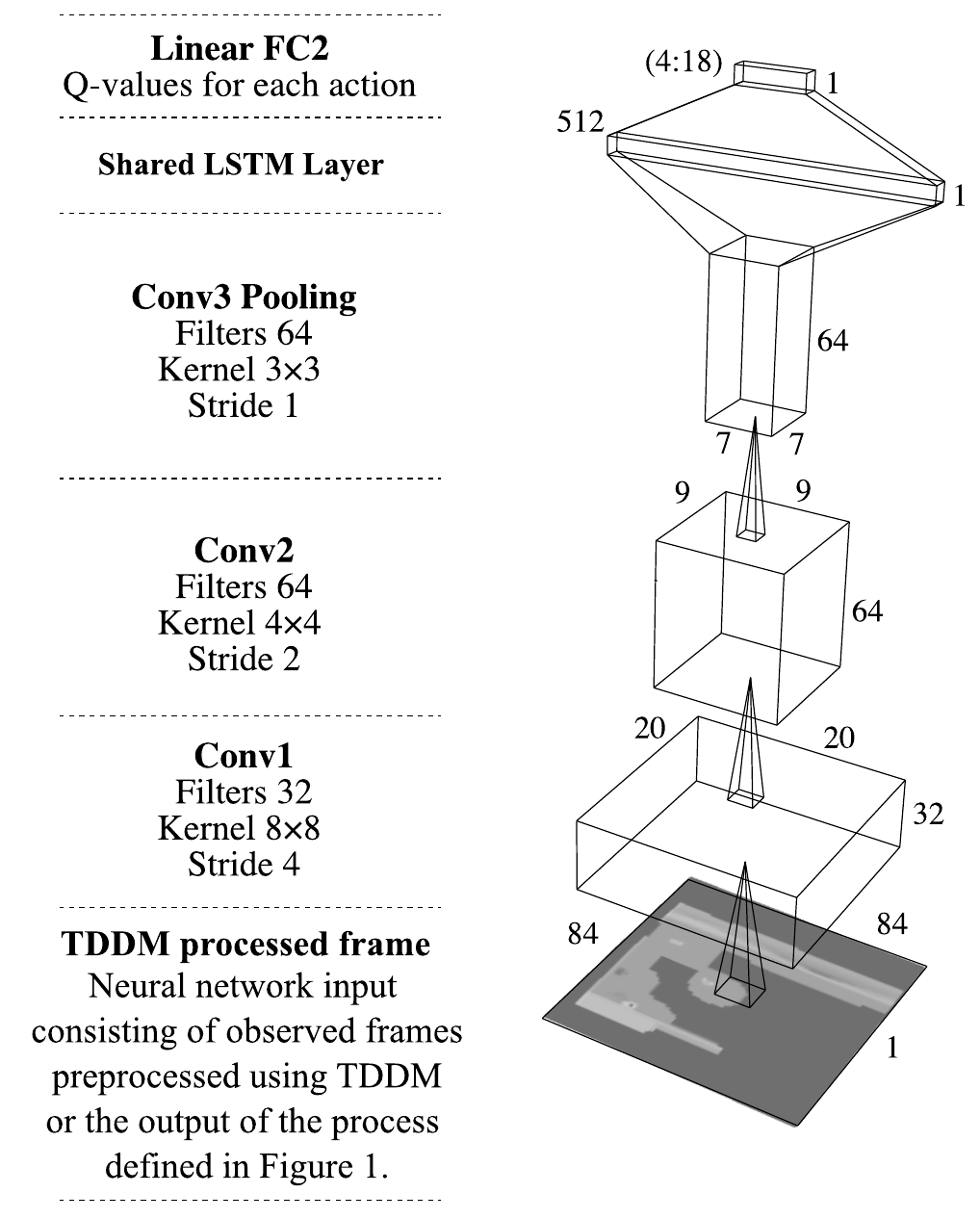}
    \caption{The final Q-value approximating component combining active state masking with three convolutional layers, a LSTM layer connected with $n-1$ sequence of previous ones and a linear fully connected one at the output. The example input is defined as a TDDM processed Atari game frame.}
    \label{fig:cnn}
\end{figure}

Figure~\ref{fig:cnn} outlines the \textit{final} component of main \textit{learning} part of the algorithm: Q-value approximation using three layers of convolutions~\cite{hausknecht2015deep} together with the proposed $TDDM$ component processing the input.

The \textit{recurrent} property of the LSTM component~\cite{hochreiter1997long} applied just before the output layer in Figure \ref{fig:cnn} is responsible for processing activations \textit{through time} allowing the ANN to infer on the transitional information from the past states. This context of previous states and actions is crucial in leaning algorithms defined over POMDP, as it provides a way to disambiguate the states of the environment. In order to achieve this, LSTM layer shown in Figure~\ref{fig:cnn} \textit{recurrently} connects with the $n$ LSTM layers that process $n$ previous agent's states. The presentation of the architecture of a LSTM is out of the scope of this work; we only mention that the basic working principle behind it is a \textit{recurrent neural network} propagation of a \textit{hidden state} $H$ through the layers. 

\begin{figure}[h!]
    \centering
    \includegraphics[width=0.66666\linewidth]{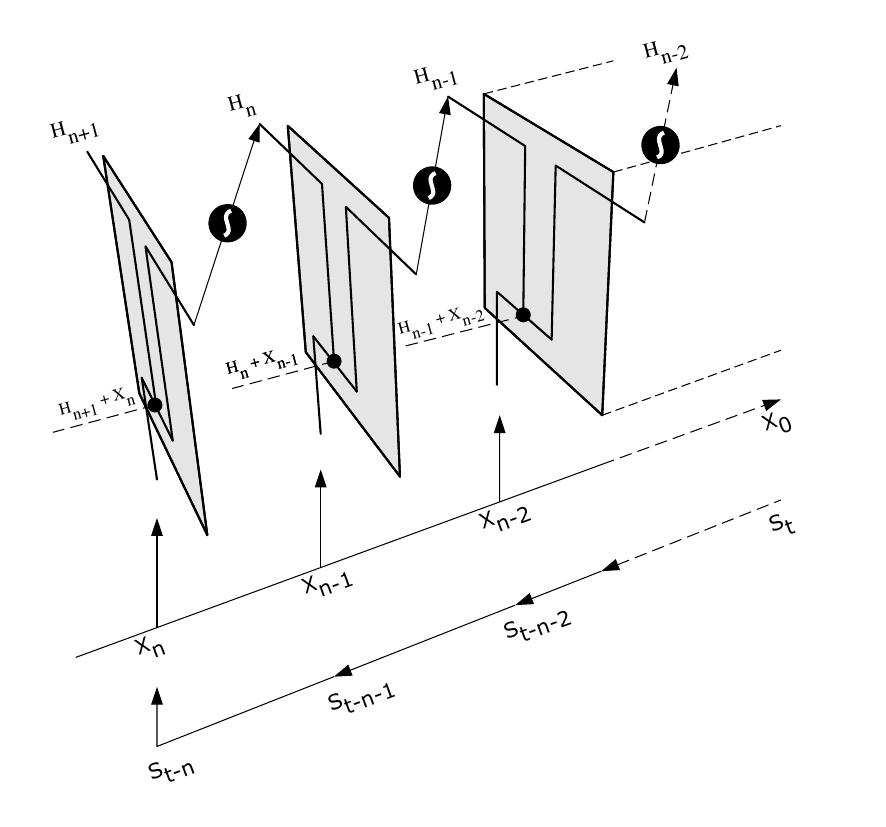}
    \caption{The \textit{hidden state} propagation of a basic recurrent neural network architecture. The inputs of the gray layers are denoted as $X$, agent's states are denoted as $S$, and hidden states by $H$. The big black circle represents the activation function applied to the layer's output, while the small one represents the \textit{concatenation} operator.}
    \label{fig:rnn}
\end{figure}

To appreciate the contribution of the LSTM \textit{recurrence} to the overall architecture, we observe a less complex RNN architecture showcased in Figure~\ref{fig:rnn}. Let's say that we want to base the agent's decision making (in our case, the approximated Q-value) not only on current perceived state, but on the $n$ previous ones, $S_t$ being the current one and $S_{t-n}$ the oldest one in our horizon. From Figure~\ref{fig:rnn} it is clear that $n$ layers are implemented, each receiving their respective temporal input $(x_n$ to $x_0)$, but at the same time each of them generating an internal \textit{hidden state} H at the \textit{output}, which becomes a part of the next layer \textit{input}, thus propagating the \textit{context} of the $n$ states.
By viewing the main architecture in this \textit{recurrent} perspective it is clear that Figure~\ref{fig:cnn} shows only the \textit{last} network out of $n$ identical ones, each being interconnected with their LSTM layers for essential recurrence property. The outlined \textit{last} layer is used to approximate the final Q-values from the outputs of the \textit{last} LSTM layer but its approximations are a product of recurrent context transfer through the previous $n-1$ LSTM layers.

\label{section:architecture}

\section{Experimental Setup}
The \textit{TDD} proof-of-concept evaluations were performed on a variety of \textit{Atari} games environments on a Python based platform mainly supported by Tensorflow~\cite{abadi2016tensorflow} and OpenAI Gym~\cite{brockman2016gym} frameworks with all of the aspects of the architecture and setup being based on the vanilla \textit{DRQN} approach originally presented by Hausknecht et al.~\cite{hausknecht2015deep}. Due to their complexity and visual variety Atari games as a set of learning problems were adopted by the original \textit{DRQN} paper~\cite{hausknecht2015deep}, we are also doing the same in our benchmarks. However, \textit{TDD} masking approach is not limited to the information that is represented in a spatial and visual way as in Atari learning problems: The \textit{temporal difference displacement} concept can be applied for selective masking of any state representations, for example a different extreme would be a flat one-dimensional vector state of a simple CartPole problem~\cite{sutton2018reinforcement}.

The purpose of the evaluation was to compare the learning performance of the \textit{baseline} \textit{DRQN}~\cite{hausknecht2015deep} with \textit{DRQN-TDDM}, an implementation that extends the \textit{baseline} to include the proposed \textit{active state masking} based on the \textit{TDDM} criterion. The \textit{DRQN} and \textit{DRQN-TDDM} implementations share the same architecture and meta-parameters;  their approximator weights and biases were randomly initialized. \textit{DRQN-TDDM} only differs in its implementation of perceptual \textit{filtering} based on a sparse \textit{TDDM} mask that is multiplied element-wise with the corresponding observations before forwarding them as an input to the learning algorithm.

Agent's policies were evaluated by performing $5$ independent learning trials for each of the two systems (DRQN and DRQN-TDDM) and averaging their achieved scores. An ANN function approximator shown in Figure~\ref{fig:cnn} was trained on each trial for a total of 7 million iterations with a \textit{root mean square propagation} (\textit{RMSProp}) optimizer capable of decaying the initial \textit{learning rate} $\alpha = 0.00025$ by a decay rate of $0.97$. The RMSProp also implemented a momentum of $0.95$ and additional gradient clipping. At each iteration the ANN's training data included a mini-batch of $64$ transitions uniformly sampled from a \textit{sliding-window} replay memory of size $800.000$.
Agent's action selection was mediated by an adjusted $\epsilon-greedy$ approach; the starting $\epsilon = 1.0$ was decayed gradually during the learning process to a final $\epsilon = 0.01$. The decay process started after the first million steps and proceeded linearly afterwards.
The discount factor $\gamma$, a parameter of the Bellman's Equation~\ref{eq:bellman-q} was set to a high value ($0.99$).

\section{Experimental Results}

\label{sec:experiments}

The results for the evaluation phase compared the learned Q-network parameters obtained in the Training phase under identical configurations. During this stage the network parameters obtained by the baseline and by $TDDM$ filtering were both evaluated with an original state input, providing a robust $TDDM$ benchmark. 

The evaluation benchmark consisted of a reproducible batch of 10 independent act-only trials for each of the ANN models obtained during the training phase. An act-only trial is characterized by acting upon the learned policy with no random exploration actions taken ($\epsilon = 0$).

Each of the independent evaluation trials were performed for a total of 100,000 steps on an original unfiltered Atari input. To guarantee reproducibility of the evaluation results a vector of 10 random scalars was generated \textit{a-priori}, specific to each game. The unique scalars have been used to seed the pseudo-random number generators of all the relevant frameworks governing the behaviour of the Atari emulator, making it deterministic with respect to the scalar used.

Faced with the identical and reproducible conditions the ANN models trained under $TDDM$ filtering outperformed the baseline ones in 20 of a total of 32 Atari game environments evaluated under the benchmark. This is reported in Table~\ref{table:results-evaluation}. The general performance measure is defined as the average return or reward that an agent received during its 10 independent batch trials; this measure represents the quantity reported along the $A.R.$ or \textit{Average Reward}, column of Table~\ref{table:results-evaluation}. Table~\ref{table:results-evaluation} outlines the summary of the performed benchmarks with the best performing values under each Atari-environment being highlighted in bold. Each row of Table~\ref{table:results-evaluation} represents an independent benchmark batch. Each game environment is represented by a total of two trial batches: the $TDDM$ and the baseline. The batches performed with the $TDDM$ models have been highlighted with a light gray background.

\begin{figure}[h!]
    \centering
    \includegraphics[width=0.66666\linewidth]{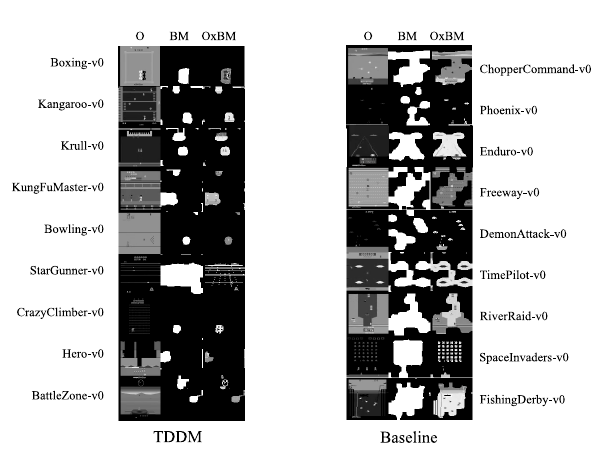}
    \caption{Masking dynamics comparing the best performing benchmarked games under TDDM models (left) and the best performing games under regular baseline models (left). Each game is represented by three frames columns, namely (O - Original unfiltered frame; BM - Binary mask obtained under $TDDM$ process outlined in \ref{fig:filter}; $OxBM$ Final filtered frame in Figure \ref{fig:filter} that is forwarded to the learning algorithm or $TDDM(S)$ created by applying element-wise matrix multiplication of $0$ with the $BM$.}
    \label{fig:maskcompare}
\end{figure}

\begin{figure}[h!]
    \centering
    \includegraphics[width=\linewidth]{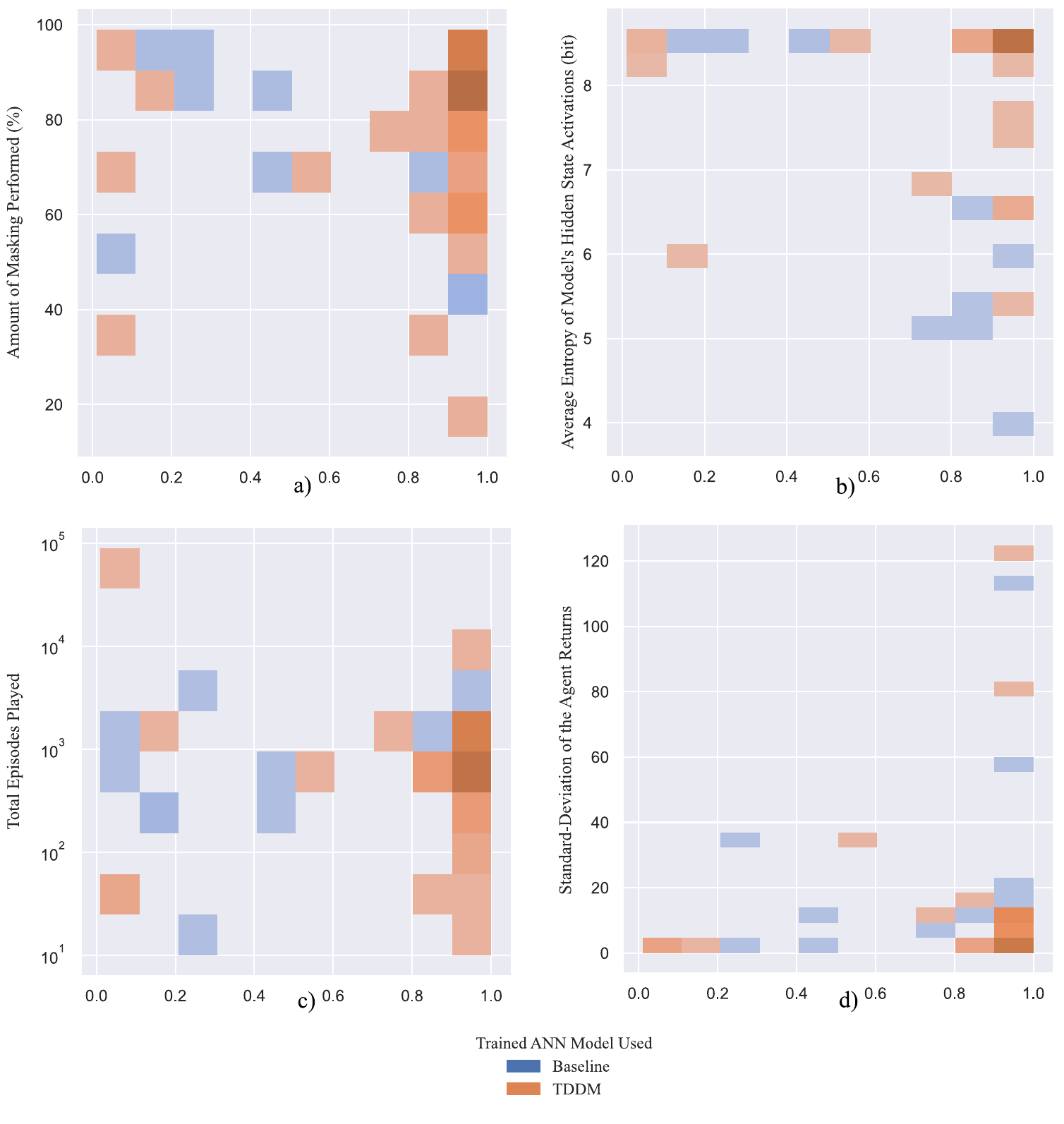}
    \caption{Visualization of distribution densities for four characterizing variables (ordinates) selected from Table~\ref{table:results-evaluation} across the relative agent's returns normalized in the range $(0 - 1)$ (abscissa). The plotted areas represent counts of ordinate-variable observations falling within each discrete bin; higher frequencies correspond to higher saturation values. Areas hues are indicative of the trained model used in the evaluation trial: benchmark results obtained using models trained under $TDDM$ approach have their frequencies or counts represented in magenta while the benchmark results obtained using baseline models are indicated in blue.}
    \label{fig:correlations}
\end{figure}

The obtained $TDDM$ masking ratios/amounts indicated in Figure~\ref{fig:correlations} a) show a very strong preference of TDDM trained models (in orange) for states that would be masked to a higher degree compared to the baseline (blue) in which this bias is not present.
The discriminatory ability of $TDDM$ models can be also appreciated from the visual comparison of the masking dynamics of the best performing games in Figure~\ref{fig:maskcompare}; Contrary to their baseline counterparts, the models trained using $TDDM$ are characterized by a much higher degree of masking, effectively removing more of the non-temporally-correlated data. Because of the $TDDM$ models ability to discriminate, the agents using $TDDM$ trained models display an artificial attention that is closer to focused attention.

For most of the evaluated games, it can be noticed that the ability of the $TDDM$ filter to discriminate between the two categories of information directly affects the performance of the $TDDM$ trained models. The cases where $TDDM$ models haven't outperformed their baseline include games that are characterised by a high amount of flickering such as \textit{DemonAttack-v0,Time-pilot-v0,Poenix-v0} and games with a high amount of repetitive synchronized movement sources such as \textit{Freeway-v0,SpaceInvaders-v0}. The low performing $TDDM$ examples as shown in the right column of Figure \ref{fig:maskcompare} held a specific set of characteristics (not limited to the above-mentioned ones) which were detrimental to the two-frame dense optical flow \cite{farneback2003two} detection accuracy and more importantly its discriminatory ability.

Although no TDDM masking was performed during the evaluation benchmark, the binary masks $BM$ were generated for analytical purposes using the identical TDD process, and are reported in Figure~\ref{fig:filter}.

TDDM models bias towards states with a higher discriminatory potential, as quantified by the amount of masking, may be seen as temporal-information greediness, or indirectly generated artificial attention capability.

This temporal-information-greedy behavior can be also observed, even more clearly, in the informational content of the LSTM states that propagate context-creating temporal information through LSTM's sequential process as depicted in Figure~\ref{fig:rnn}. 

In order to quantify the ability of $TDDM$ models to distinguish between the static and temporally correlated information, and then focus on the latter, it is possible to examine the actual level of utilization of the LSTM \textit{recurrent} layer during the realization of the agent's policy. The LSTM utilization is represented by the amount of informational entropy contained in the layer's hidden states, which, in case of LSTM architectures, are the main propagators of contextual, temporally related information. The experimental measures reported in section $b)$ of Figure~\ref{fig:correlations} indicate that, regardless of the presented environment/game, the $TDDM$ learned models displayed levels of their $LSTM$ layer hidden states entropy significantly higher than their baseline counterparts. This trend is visible in the upper right corner of the plot showing a high distribution density of the $TDDM$ category.

As this type of information is more crucial in forming the agent's belief state, the agents that exhibit bias towards it can be seen as more effective in their utilization of the communication channel between their own perception and the learning algorithm. The agents that used $TDDM$ models in identical and reproducible benchmark trials, according to Figure~\ref{fig:correlations} $b)$ propagated higher levels of temporally correlated information from the environment to the learning algorithm.

The ability of (artificial or biological) agent to convey a specific type of information that is more descriptive of the environment dynamics, or perturbations, increases its ability to produce more credible belief states. As we can see from Figure~\ref{fig:correlations} $(b)$ acting optimally in a dynamic environment benefits from belief state representations that indeed contain in themselves necessary temporal abstractions crucial to organism survival and, in the case of an artificial agent, to its ability to maximize the expected return of a reinforcement function in the long run.

While the a) and b) plots of Figure~\ref{fig:correlations} are mostly descriptive of the difference in information processing dynamics, the second row (plots c) and d)) puts in evidence the difference in exploration/exploitation dispositions of the agents using models trained with $TDDM$ approach with respect to the baseline ones. From Figure~\ref{fig:correlations} c) it is evident that the $TDDM$ models have produced policies that in general allow for longer Atari game episodes, which, in most of the game variations, accounts for a higher exploration rate of the game state space.

On the contrary, Figure~\ref{fig:correlations} d)  indicates that rewards for the $TDDM$ models are more consistent, as quantified by their variance. While plot~\ref{fig:correlations} c) seems to suggest a more efficient exploration of the game state space, the d) plot also accounts for the $TDDM$ models ability to exploit the reliability of Q-value predictions in such a way as to be able to predict the return more consistently than their baseline counterparts.

\begin{table}[h!]
\centering
\tiny
\begin{tabular}{ l|l|l|l|l|l|l|l|l|l|l } 
Environment & A.R & N.P.E & H.S.A.E & S.A.E & H.S.S & M.A & S.S & ST.D.R & ST.D.A & ST.D.M \\
\hline
Alien-v0 & 0.7064 & \B 747 & 5.094 & \B 8.56 & 1.166 & \B 70.42 & 214.1 & 4.073 & \B 6.094 & \B 700.5 \\
\rowcolor{gray!7} Alien-v0 & \B 0.7889 & 670.5 & \B 6.664 & 8.531 & \B 2.423 & 70.03 & \B 217 & \B 8.127 & 4.734 & 656.7 \\
Asterix-v0 & 0.6217 & 1280 & 4.981 & \B 8.437 & 1.769 & 74.09 & \B 382.3 & 5.56 & 1.388 & \B 1095 \\
\rowcolor{gray!7} Asterix-v0 & \B 0.7964 & \B 1366 & \B 7.35 & 8.333 & \B 2.657 & \B 82.03 & 315.5 & \B 6.286 & \B 1.997 & 1031 \\
Asteroids-v0 & \B 1.436 & \B 1165 & 8.61 & 8.61 & \B 10.24 & 96.78 & \B 512 & 9.586 & \B 3.753 & 191.1 \\
\rowcolor{gray!7} Asteroids-v0 & 1.349 & 792.9 & \B 8.61 & \B 8.61 & 10.24 & \B 96.79 & 512 & \B 9.641 & 3.573 & \B 208.8 \\
Atlantis-v0 & 9.158 & 649.1 & 6.652 & 8.472 & 2.71 & \B 78.61 & 359.7 & 111.6 & 1.197 & 694 \\
\rowcolor{gray!7} Atlantis-v0 & \B 9.805 & \B 675.5 & \B 8.48 & \B 8.582 & \B 6.105 & 77.78 & \B 416.6 & \B 124.7 & \B 1.293 & \B 699.7 \\
BattleZone-v0 & 0.664 & \B 2561 & \B 8.61 & 8.61 & \B 10.24 & \B 84.22 & \B 512 & 32.67 & \B 3.785 & \B 1115 \\
\rowcolor{gray!7} BattleZone-v0 & \B 2.293 & 566.7 & 8.61 & \B 8.61 & \B 10.24 & 52.74 & \B 512 & \B 81.9 & 1.181 & 752.4 \\
BeamRider-v0 & 0.05842 & \B 8.176e+04 & \B 8.61 & \B 8.61 & \B 10.24 & \B 92.4 & \B 512 & 1.627 & 0.9999 & \B 1220 \\
\rowcolor{gray!7} BeamRider-v0 & \B 0.737 & 215.2 & 8.608 & 8.599 & 10.02 & 13.14 & 510 & \B 5.744 & \B 2.062 & 1100 \\
Berzerk-v0 & \B 1.063 & \B 2066 & \B 8.6 & \B 8.6 & \B 10.05 & 73.34 & \B 510.7 & \B 7.256 & \B 4.509 & \B 959.8 \\
\rowcolor{gray!7} Berzerk-v0 & 0.9837 & 1958 & 8.599 & 8.566 & 8.988 & \B 73.74 & 500 & 6.973 & 4.366 & 947 \\
Bowling-v0 & 0.000845 & 10 & \B 8.61 & \B 8.61 & \B 10.24 & 98.1 & \B 512 & 0.07489 & 0.8484 & \B 180.3 \\
\rowcolor{gray!7} Bowling-v0 & \B 0.002796 & \B 18.6 & 8.61 & 8.61 & \B 10.24 & \B 98.99 & \B 512 & \B 0.09414 & \B 0.8768 & 121.5 \\
Boxing-v0 & -0.008385 & 57.9 & 5.35 & 6.723 & 1.394 & 89.67 & 125.1 & \B 0.3677 & 3.736 & 450.4 \\
\rowcolor{gray!7} Boxing-v0 & \B 0.004923 & \B 65.9 & \B 7.714 & \B 8.616 & \B 6.63 & \B 91.44 & \B 388.2 & 0.3479 & \B 3.999 & \B 471.4 \\
Breakout-v0 & 0.00021 & \B 8.958e+04 & \B 8.61 & 8.61 & \B 10.24 & 48.97 & \B 512 & 0.01679 & \B 0.9616 & \B 1501 \\
\rowcolor{gray!7} Breakout-v0 & \B 0.02164 & 1225 & 8.61 & \B 8.61 & 10.24 & \B 92.97 & \B 512 & \B 0.1991 & 0.9374 & 326 \\
ChopperCommand-v0 & \B 1.675 & \B 526.6 & \B 8.602 & \B 8.598 & \B 9.271 & \B 60.17 & \B 510.7 & \B 14.9 & \B 5.761 & \B 713.9 \\
\rowcolor{gray!7} ChopperCommand-v0 & 1.39 & 459.4 & 8.591 & 8.594 & 4.14 & 56.91 & 427.6 & 14.64 & 2.521 & 623.4 \\
CrazyClimber-v0 & 0.1476 & \B 283.2 & 8.571 & \B 8.608 & \B 10.18 & 84.66 & \B 511.9 & 3.839 & 0.6595 & 538.5 \\
\rowcolor{gray!7} CrazyClimber-v0 & \B 1.103 & 157 & \B 8.612 & 8.567 & 6.56 & \B 89.08 & 482.6 & \B 10.45 & \B 1.763 & \B 784.3 \\
DemonAttack-v0 & \B 0.2449 & 819.1 & 8.61 & 8.61 & 10.1 & 87.64 & \B 512 & \B 1.702 & 1.519 & \B 763.3 \\
\rowcolor{gray!7} DemonAttack-v0 & 0.004 & \B 5.231e+04 & \B 8.61 & \B 8.61 & \B 10.23 & \B 92.12 & 512 & 0.2015 & \B 1.646 & 428.3 \\
Enduro-v0 & \B 0.02173 & \B 30 & \B 8.125 & 8.61 & 6.046 & \B 70.53 & 467.1 & \B 0.2403 & \B 2.655 & \B 1148 \\
\rowcolor{gray!7} Enduro-v0 & 0.001997 & \B 30 & 8.116 & \B 8.612 & \B 9.018 & 66.62 & \B 482.7 & 0.1799 & 2.625 & 1082 \\
FishingDerby-v0 & \B -0.02089 & 52.8 & \B 8.61 & 8.61 & \B 10.24 & \B 40.48 & \B 512 & \B 0.277 & 5.345 & 315.3 \\
\rowcolor{gray!7} FishingDerby-v0 & -0.03175 & \B 53.6 & 8.61 & \B 8.61 & 10.24 & 38.22 & \B 512 & 0.2686 & \B 5.455 & \B 334.7 \\
Freeway-v0 & 0 & \B 50.1 & \B 8.61 & \B 8.61 & 10.24 & \B 39.99 & 512 & 0 & 0.1281 & \B 664.8 \\
\rowcolor{gray!7} Freeway-v0 & \B 0.009295 & 50 & 8.61 & 8.61 & \B 10.24 & 37.37 & \B 512 & \B 0.09596 & \B 0.5826 & 663.2 \\
Frostbite-v0 & 0.2432 & \B 4452 & 5.831 & 8.475 & 1.842 & 57.26 & 227.6 & 1.54 & \B 5.08 & \B 1897 \\
\rowcolor{gray!7} Frostbite-v0 & \B 0.2599 & 1249 & \B 6.526 & \B 8.482 & \B 3.041 & \B 59.3 & \B 270.7 & \B 1.591 & 4.58 & 1817 \\
Hero-v0 & 0.02497 & 343.9 & \B 8.492 & 8.521 & \B 3.952 & \B 91.08 & \B 388.9 & 2.697 & 3.828 & \B 724.5 \\
\rowcolor{gray!7} Hero-v0 & \B 0.1651 & \B 473.4 & 5.267 & \B 8.568 & 3.383 & 90.76 & 313.9 & \B 8.428 & \B 5.199 & 682.4 \\
IceHockey-v0 & -0.0048 & \B 28.6 & 8.61 & \B 8.61 & \B 10.24 & 87.59 & \B 512 & 0.07606 & 3.889 & 277.8 \\
\rowcolor{gray!7} IceHockey-v0 & \B -0.002695 & 27 & \B 8.61 & 8.61 & \B 10.24 & \B 88.08 & \B 512 & \B 0.08236 & \B 3.934 & \B 308.5 \\
Jamesbond-v0 & 0.2138 & 932.4 & \B 8.61 & \B 8.61 & \B 10.24 & \B 60.55 & \B 512 & 3.263 & \B 5.016 & 1098 \\
\rowcolor{gray!7} Jamesbond-v0 & \B 0.2576 & \B 1188 & 8.603 & 8.601 & 10.14 & 59.98 & 511.6 & \B 3.58 & 4.882 & \B 1110 \\
Kangaroo-v0 & 0.4452 & \B 865.4 & \B 8.603 & \B 8.603 & \B 10.1 & 83.45 & \B 511.4 & 9.43 & \B 4.22 & \B 768.8 \\
\rowcolor{gray!7} Kangaroo-v0 & \B 0.9237 & 827.7 & 8.586 & 8.578 & 10.1 & \B 83.73 & 510.6 & \B 13.56 & 3.765 & 757.4 \\
Krull-v0 & 0.2896 & 161.9 & \B 8.516 & 8.576 & 5.926 & \B 68.56 & 425.3 & 1.901 & 4.28 & 2001 \\
\rowcolor{gray!7} Krull-v0 & \B 0.6939 & \B 176.1 & 8.146 & \B 8.61 & \B 7.116 & 61.28 & \B 432.6 & \B 2.919 & \B 5.47 & \B 2299 \\
KungFuMaster-v0 & 0.0282 & \B 987 & \B 8.61 & \B 8.61 & \B 10.24 & \B 95.64 & \B 512 & 2.375 & 3.216 & 401.1 \\
\rowcolor{gray!7} KungFuMaster-v0 & \B 0.7838 & 698.4 & 8.567 & 8.532 & 3.271 & 87.42 & 376.8 & \B 12 & \B 4.123 & \B 687.7 \\
Phoenix-v0 & \B 0.6441 & 1255 & 3.841 & 8.63 & 0.7936 & 83.61 & 223.3 & \B 7.617 & \B 2.068 & \B 636 \\
\rowcolor{gray!7} Phoenix-v0 & 0.133 & \B 1496 & \B 6.025 & \B 8.632 & \B 1.855 & \B 86.67 & \B 366.4 & 2.776 & 0.6282 & 487.4 \\
Pitfall-v0 & -0.05164 & 1273 & \B 8.61 & \B 8.61 & \B 10.24 & \B 82.39 & \B 512 & \B 0.4363 & 1.865 & \B 876.9 \\
\rowcolor{gray!7} Pitfall-v0 & \B -0.05085 & \B 1.332e+04 & 8.61 & 8.61 & 10.24 & 80.52 & 512 & 0.4312 & \B 5.258 & 710.4 \\
Pong-v0 & \B -0.004044 & 35.4 & 8.61 & \B 8.61 & 10.24 & \B 91.32 & 512 & 0.1014 & \B 1.803 & \B 429.7 \\
\rowcolor{gray!7} Pong-v0 & -0.01118 & \B 68.5 & \B 8.61 & 8.61 & \B 10.24 & 89.4 & \B 512 & \B 0.1271 & 1.524 & 412.1 \\
Qbert-v0 & 1.367 & 1615 & \B 6.482 & \B 8.478 & 2.579 & \B 84.69 & \B 294.8 & 12.98 & \B 1.729 & 300.2 \\
\rowcolor{gray!7} Qbert-v0 & \B 1.599 & \B 1964 & 6.477 & 8.469 & \B 2.922 & 83.97 & 284.2 & \B 13.77 & 1.494 & \B 305.9 \\
Riverraid-v0 & \B 3.015 & 880.9 & \B 7.577 & \B 8.451 & \B 3.429 & \B 74.52 & 323 & \B 22.01 & \B 6.135 & \B 1302 \\
\rowcolor{gray!7} Riverraid-v0 & 2.349 & \B 1613 & 6.701 & 8.323 & 2.776 & 74.42 & \B 329.6 & 13.38 & 5.081 & 1261 \\
Seaquest-v0 & \B 0.3839 & 627 & \B 8.61 & \B 8.61 & \B 10.21 & \B 75.03 & \B 511.9 & \B 2.764 & \B 4.838 & \B 748.6 \\
\rowcolor{gray!7} Seaquest-v0 & 0.3393 & \B 682.9 & 8.588 & 8.577 & 9.127 & 73.99 & 506.7 & 2.604 & 4.687 & 720.5 \\
SpaceInvaders-v0 & \B 0.4755 & 761.7 & 8.499 & \B 8.607 & \B 8.431 & \B 82.9 & \B 504.5 & \B 3.874 & \B 1.593 & 1017 \\
\rowcolor{gray!7} SpaceInvaders-v0 & 0.3813 & \B 918 & \B 8.506 & 8.598 & 7.273 & 82.1 & 493.2 & 3.576 & 1.561 & \B 1073 \\
StarGunner-v0 & 0.0472 & 484.3 & \B 8.61 & \B 8.61 & \B 10.24 & 71.69 & \B 512 & 2.901 & 1.535 & \B 743.2 \\
\rowcolor{gray!7} StarGunner-v0 & \B 0.7814 & \B 721.6 & 8.61 & 8.61 & \B 10.24 & \B 72.44 & \B 512 & \B 12.47 & \B 2.567 & 699.2 \\
TimePilot-v0 & \B 2.172 & \B 815.5 & 8.61 & 8.61 & \B 10.22 & \B 75.96 & \B 511.8 & \B 58.96 & \B 2.132 & 1329 \\
\rowcolor{gray!7} TimePilot-v0 & 1.132 & 779.3 & \B 8.674 & \B 8.637 & 2.69 & 70.23 & 338.4 & 34.64 & 1.27 & \B 1344 \\
\hline
\rowcolor{gray!10} \B \#TDDM & \B 20 & 18 & 13 & 13 & 16 & 19 & 16 & 20 & 15 & 12 \\
\hline
\rowcolor{gray!10
} \B \#benchmark & \B 12 & 15 & 19 & 19 & 20 & 13 & 22 & 12 & 17 & 20 \\
\hline

\end{tabular}
\caption{Results of the Evaluation Benchmark performed under identical reproducible setups. Best performing batches are outlined in bold for each of the game environments and results obtained with models trained under $TDDM$ are highlighted with light-gray background; The last two rows represent a summary of best performing values for each of the columns: \#TDDM row represents the number of best values among the batches obtained by using models trained under $TDDM$ while the \#Benchmark row does the same with the baseline models. Detailed description of the specific columns used are presented in Table~\ref{table:description}.}
\label{table:results-evaluation}
\end{table}

\begin{table}[h!]
    \centering
    \tiny
    \begin{tabular}{l|l|m{7.8cm}}
\B Abbreviation & \B Full Name & \B Description\\[0.5cm]
Environment & Environment & Specific Atari game used in benchmark batch. \\[0.5cm]
A.R. & Average Return & The immediate rewards that the agents received averaged over all of the 10 trials that formed a single benchmark batch. \\[0.5cm]
N.P.E. & Number of Played Episodes & Average Total Number of Played Episodes in a single trial. \\[0.5cm]
H.S.A.E. & Hidden States Activation Entropy & Average Shannon's entropy in bits of the model's hidden states $H_n$ indicative of the amount of information being effectively propagated through their activations in the LSTM part of the main ANN model detailed in \ref{fig:rnn}. \\[0.5cm]
S.A.E. & States Activation Entropy & Average Shannon's entropy in bits of the model's input states $X_n$ indicative of the amount of information being effectively propagated through their activations in the LSTM part of the main ANN model detailed in \ref{fig:rnn}. \\[0.5cm]
H.S.S. & Hidden States Sparsity & Average Percentage of Non-Zero Hidden States Activations. Higher percentage indicates more activity in RNN Hidden States propagation. \\[0.5cm]
M.A. & Masking Amount & Percentage of the input state's pixels masked or blanked with $TDDM$. \\[0.5cm]
S.S. & States Sparsity & Percentage of Non-Zero model's input states $X_n$ Activations. Higher percentage indicates more activity in RNN input state propagation. \\[0.5cm]
ST.D.R & Standard-Deviation of Returns & Depending on a specific environment reinforcement function the variance of the Returns could be an indicative of an agent's preference of exploration over exploitation. \\[0.5cm]
ST.D.A & Standard-Deviation of Selected Actions & Depending on a specific environment configuration the variance of the Actions taken could be an indicative of an agent's preference of exploration over exploitation. \\[0.5cm]
ST.D.M & Standard-Deviation of Masking Percentages & The variance in Masking Amounts of single frames could indicate the level of adaptability of the motion detection technique shown in Figure~\ref{fig:filter} to a specific environment.
    \end{tabular}
    \caption{Detailed Description of the type of data represented in the columns of Table~\ref{table:results-evaluation}.
    }
    \label{table:description}
\end{table}

\section{Concluding Remarks}
The abundance of the inherent information-generating \textit{uncertainty} in our perception of the world pushed the human evolution into a momentum of producing information-processing mechanisms with increasing complexity that would in turn be able to reduce this uncertainty on a variety of levels or abstraction including crafting our immediate environment by creating patterns of predictability; be it in a form of ubiquitous technical systems (which include the artificial learning ones presented in this work) or the more abstract social structures.

\textit{Interaction} and \textit{causal} relationships with the perceived environment are emerging as the focus of the perception-based information-gathering process rather than expanding the perception domain itself. For example, an agent would not benefit from a hypothetical super-perception that would enable it to perceive the amount of information contained in the spin direction of electrons in each of the atoms of a typical physical object; if a state of spin can be either spin-up or spin-down with equal probability distribution, this would give us an entropy of 1 bit per electron. This information though, will not support the determination of the interaction that physical objects have with their surroundings or provide a learning mechanism with the representation of perturbations of the environment significant to its survival, e.g., temperature, sound, vision.

Interacting with a more predictable environment reduces the overall information that needs to be processed, but at the same time generates more information that explains the \textit{causal} relationships within.
This \textit{causal} subset of the perceived information can be seen as an information \textit{gain} of interaction (or transition from $s$ to $s'$ in our case) that is irreducible to its composing parts ($s$ and $s'$) according to the integrated information principle or $\Phi$ proposed by~\cite{tononi2016integrated}.

Depending on the specific machine learning approach, an artificial learning agent capable of discriminating the perceived information based on a temporal or a spatial context can be more effective to convert that same information into higher order representations, such as Q-values that would eventually lead to the creation of better policies.

The main insight obtained from this work is that perceptual discrimination based on temporal difference displacement, or $TDD$ criterion, as evident from Table \ref{table:results-evaluation}, may enable convergence of temporal-difference learning algorithms to their optimal policies in fewer learning steps
; moreover, it can produce learned models that perform better than a state-of-art baseline in 20 out of 32 different Atari games, under the identical and reproducible setups.

It can be also noted that the models learned under $TDD$ masking possess a strong tendency towards an increased utilization of the recurrent LSTM section of the main Q-approximator shown in Figure~\ref{fig:cnn}, effectively utilizing more of the temporally correlated information (\cite{tononi2016integrated}) in the creation of the agent's belief state leading to overall better agent's performances in the benchmark, as indicated in Table~\ref{table:results-evaluation}.

\section{Acknowledgments}
This article has been supported by the OP RDE funded project Research Center for Informatics No.: CZ.02.1.01/0.0./0.0./16\_019/0000765.

\appendix
\section*{Appendix A.}
\label{app:training}

In this appendix we present the variation of a total of three variables characterizing the actual learning during the agent's training phase, namely: total cumulative return, average Q-value, and number of played episodes.

\begin{table*}[h!]
\centering
\tiny
\begin{tabular}{ |l|l|l|l|l|l|l| } 

Environment & Mask & Mean & Maximum & Minimum & Median & Mod \\
Alien-v0 & 0 & 0.7126 & 1.352 & 0.232 & 0.714 & 0.714 \\
\rowcolor{gray!7} \B Alien-v0 & \B 1 & \B 0.801 & \B 1.536 & \B 0.258 & \B 0.814 & \B 0.75 \\
Asterix-v0 & 0 & 1.182 & \B 5.57 & \B 0.72 & 1.19 & 1.21 \\
\rowcolor{gray!7} \B Asterix-v0 & \B 1 & \B 1.654 & 5.07 & 0.71 & \B 1.74 & \B 1.74 \\
Asteroids-v0 & 0 & 0.781 & \B 4.274 & 0.21 & 0.764 & 0.524 \\
\rowcolor{gray!7} \B Asteroids-v0 & \B 1 & \B 0.9593 & 4.196 & \B 0.312 & \B 0.964 & \B 0.988 \\
\B Atlantis-v0 & \B 0 & \B 12.29 & \B 70.16 & \B 5.62 & \B 12.3 & \B 12.08 \\
\rowcolor{gray!7} Atlantis-v0 & 1 & 10.7 & 65.78 & 5.6 & 10.58 & 11.42 \\
BattleZone-v0 & 0 & 0.3694 & 12.4 & \B 0 & \B 0 & \B 0 \\
\rowcolor{gray!7} \B BattleZone-v0 & \B 1 & \B 0.4346 & \B 13 & \B 0 & \B 0 & \B 0 \\
BeamRider-v0 & 0 & 0.4709 & \B 1.461 & 0.0528 & 0.4776 & \B 0.3168 \\
\rowcolor{gray!7} \B BeamRider-v0 & \B 1 & \B 0.5124 & 1.302 & \B 0.132 & \B 0.5564 & 0.2464 \\
\B Berzerk-v0 & \B 0 & \B 0.8657 & 2.94 & 0.39 & \B 0.89 & \B 0.91 \\
\rowcolor{gray!7} Berzerk-v0 & 1 & 0.8321 & \B 3.44 & \B 0.42 & 0.84 & 0.82 \\
\B Bowling-v0 & \B 0 & \B 0.01267 & 0.0464 & \B 0 & \B 0.0118 & \B 0.012 \\
\rowcolor{gray!7} Bowling-v0 & 1 & 0.009806 & \B 0.0468 & \B 0 & 0.0096 & 0.0092 \\
Boxing-v0 & 0 & 0.00288 & 0.024 & -0.0202 & 0.0026 & 0.0002 \\
\rowcolor{gray!7} \B Boxing-v0 & \B 1 & \B 0.01691 & \B 0.0458 & \B -0.0058 & \B 0.0176 & \B 0.0134 \\
Breakout-v0 & 0 & 0.01451 & 0.0356 & 0.0022 & 0.0152 & 0.0158 \\
\rowcolor{gray!7} \B Breakout-v0 & \B 1 & \B 0.02939 & \B 0.0604 & \B 0.0032 & \B 0.0312 & \B 0.032 \\
ChopperCommand-v0 & 0 & 1.322 & \B 3.82 & 0 & 1.28 & 1.18 \\
\rowcolor{gray!7} \B ChopperCommand-v0 & \B 1 & \B 1.796 & 3.6 & \B 0.14 & \B 1.84 & \B 1.9 \\
CrazyClimber-v0 & 0 & 1.789 & 13.18 & \B 0.4 & 1.68 & 1.68 \\
\rowcolor{gray!7} \B CrazyClimber-v0 & \B 1 & \B 3.055 & \B 13.7 & 0.3 & \B 2.78 & \B 2.28 \\
\B DemonAttack-v0 & \B 0 & \B 0.2643 & \B 1.014 & 0.022 & \B 0.242 & \B 0.242 \\
\rowcolor{gray!7} DemonAttack-v0 & 1 & 0.1906 & 0.827 & \B 0.053 & 0.1845 & 0.177 \\
\B Enduro-v0 & \B 0 & \B 0.02432 & \B 0.0672 & \B -0.0024 & \B 0.0246 & \B 0 \\
\rowcolor{gray!7} Enduro-v0 & 1 & 0.02033 & 0.065 & -0.0032 & 0.02 & \B 0 \\
\B FishingDerby-v0 & \B 0 & \B -0.02784 & \B -0.002 & -0.241 & \B -0.0246 & \B -0.0176 \\
\rowcolor{gray!7} FishingDerby-v0 & 1 & -0.03838 & -0.0184 & \B -0.2332 & -0.0376 & -0.0344 \\
Freeway-v0 & 0 & 0 & 0 & \B 0 & 0 & \B 0 \\
\rowcolor{gray!7} \B Freeway-v0 & \B 1 & \B 0.006123 & \B 0.0118 & \B 0 & \B 0.007 & \B 0 \\
\B Frostbite-v0 & \B 0 & \B 0.3101 & \B 0.968 & 0.106 & \B 0.314 & \B 0.352 \\
\rowcolor{gray!7} Frostbite-v0 & 1 & 0.2978 & 0.844 & \B 0.144 & 0.296 & 0.288 \\
IceHockey-v0 & 0 & -0.003464 & \B 0.0004 & \B -0.0144 & -0.0036 & -0.004 \\
\rowcolor{gray!7} \B IceHockey-v0 & \B 1 & \B -0.002848 & \B 0.0004 & -0.0172 & \B -0.0028 & \B -0.0028 \\
\B Jamesbond-v0 & \B 0 & \B 0.1498 & \B 0.31 & 0 & \B 0.17 & \B 0.2 \\
\rowcolor{gray!7} Jamesbond-v0 & 1 & 0.1491 & 0.29 & \B 0.01 & 0.16 & 0.17 \\
Kangaroo-v0 & 0 & 0.5308 & 1.2 & \B 0 & 0.56 & 0.68 \\
\rowcolor{gray!7} \B Kangaroo-v0 & \B 1 & \B 0.7284 & \B 1.48 & \B 0 & \B 0.76 & \B 0.72 \\
\B Krull-v0 & \B 0 & \B 1.174 & 5.649 & \B 0.2232 & \B 1.176 & \B 1.397 \\
\rowcolor{gray!7} Krull-v0 & 1 & 1.122 & \B 6.226 & 0.045 & 1.103 & 0.8262 \\
KungFuMaster-v0 & 0 & 0.1157 & 3.28 & 0 & 0.06 & 0 \\
\rowcolor{gray!7} \B KungFuMaster-v0 & \B 1 & \B 1.911 & \B 4.88 & \B 0.36 & \B 1.88 & \B 1.92 \\
\B Pitfall-v0 & \B 0 & \B -0.02355 & \B 0 & \B -0.4292 & \B 0 & \B 0 \\
\rowcolor{gray!7} Pitfall-v0 & 1 & -0.04863 & \B 0 & -0.4698 & -0.0366 & \B 0 \\
\B Pong-v0 & \B 0 & \B -0.009172 & \B 0.001 & \B -0.102 & \B -0.0052 & \B -0.004 \\
\rowcolor{gray!7} Pong-v0 & 1 & -0.01017 & -0.0008 & -0.1034 & -0.0094 & -0.0056 \\
Qbert-v0 & 0 & 0.7171 & 2.15 & 0.2 & 0.71 & 0.68 \\
\rowcolor{gray!7} \B Qbert-v0 & \B 1 & \B 0.8616 & \B 2.185 & \B 0.255 & \B 0.865 & \B 0.845 \\
Riverraid-v0 & 0 & 3.017 & 10.46 & \B 1.62 & 3.008 & 2.636 \\
\rowcolor{gray!7} \B Riverraid-v0 & \B 1 & \B 3.033 & \B 10.79 & 1.304 & \B 3.085 & \B 3.19 \\
Seaquest-v0 & 0 & 0.2961 & 0.768 & 0.104 & 0.3 & 0.3 \\
\rowcolor{gray!7} \B Seaquest-v0 & \B 1 & \B 0.3279 & \B 0.872 & \B 0.12 & \B 0.332 & \B 0.344 \\
\B SpaceInvaders-v0 & \B 0 & \B 0.3921 & 1.519 & \B 0.21 & \B 0.392 & \B 0.394 \\
\rowcolor{gray!7} SpaceInvaders-v0 & 1 & 0.3533 & \B 1.525 & 0.161 & 0.35 & 0.363 \\
StarGunner-v0 & 0 & 0.4757 & 2.56 & \B 0.1 & 0.48 & 0.48 \\
\rowcolor{gray!7} \B StarGunner-v0 & \B 1 & \B 0.5605 & \B 3.06 & 0.08 & \B 0.54 & \B 0.52 \\
\B TimePilot-v0 & \B 0 & \B 1.688 & \B 11.32 & 0.28 & \B 1.72 & 1.16 \\
\rowcolor{gray!7} TimePilot-v0 & 1 & 1.403 & 9.36 & \B 0.38 & 1.28 & \B 1.24 \\

\end{tabular}
\caption{Total cumulative return received during the training phase for each of the combinations of environment/masking. Best values are highlighted in bold.}
\label{table:results-reward}
\end{table*}

\begin{table*}[h!]
\centering
\tiny
\begin{tabular}{ |l|l|l|l|l|l|l| } 

Environment & Mask & Mean & Maximum & Minimum & Median & Mod \\
Alien-v0 & 0 & 232.6 & 363.8 & \B 0.1028 & 259.5 & 0.1028 \\
\rowcolor{gray!7} \B Alien-v0 & \B 1 & \B 419.8 & \B 577 & 0.1018 & \B 459.5 & \B 443.9 \\
Asterix-v0 & 0 & 210.6 & 336.6 & \B 0.09456 & 239 & \B 0.09456 \\
\rowcolor{gray!7} \B Asterix-v0 & \B 1 & \B 481.2 & \B 741.5 & 0.08503 & \B 646.7 & 0.08503 \\
Asteroids-v0 & 0 & 43.38 & \B 171.4 & \B 0.04762 & 29.72 & \B 0.04762 \\
\rowcolor{gray!7} \B Asteroids-v0 & \B 1 & \B 56.83 & 118.5 & 0.04545 & \B 42.03 & 0.04545 \\
Atlantis-v0 & 0 & 1357 & 2138 & 0.08246 & 1689 & 0.08246 \\
\rowcolor{gray!7} \B Atlantis-v0 & \B 1 & \B 1906 & \B 2639 & \B 0.08319 & \B 2342 & \B 2476 \\
BattleZone-v0 & 0 & 0.6521 & 1.45 & \B 0.01402 & 0.6103 & \B 0.01402 \\
\rowcolor{gray!7} \B BattleZone-v0 & \B 1 & \B 0.721 & \B 1.595 & 0.01006 & \B 0.7024 & 0.01006 \\
BeamRider-v0 & 0 & 179.2 & 409.1 & \B 0.0283 & 147.3 & \B 0.0283 \\
\rowcolor{gray!7} \B BeamRider-v0 & \B 1 & \B 347.1 & \B 543.9 & 0.02286 & \B 453.2 & 0.02286 \\
\B Berzerk-v0 & \B 0 & \B 206.8 & \B 314.1 & 0.04415 & \B 276.8 & \B 283.7 \\
\rowcolor{gray!7} Berzerk-v0 & 1 & 168.6 & 233.6 & \B 0.05318 & 200.3 & 0.05318 \\
Bowling-v0 & 0 & 4.137 & 11.47 & \B 0.02746 & 1.88 & \B 0.02746 \\
\rowcolor{gray!7} \B Bowling-v0 & \B 1 & \B 8.175 & \B 16.58 & 0.01117 & \B 8.002 & 0.01117 \\
Boxing-v0 & 0 & 139.5 & 279.8 & \B 0.01625 & 132.4 & \B 0.01625 \\
\rowcolor{gray!7} \B Boxing-v0 & \B 1 & \B 183 & \B 305.4 & 0.006279 & \B 203.6 & 0.006279 \\
Breakout-v0 & 0 & 8.023 & 11.11 & \B 0.04226 & 9.65 & 9.674 \\
\rowcolor{gray!7} \B Breakout-v0 & \B 1 & \B 27.91 & \B 38.4 & 0.03465 & \B 29.45 & \B 28.56 \\
ChopperCommand-v0 & 0 & 185.3 & 503.7 & 0.02374 & 66.08 & 0.02374 \\
\rowcolor{gray!7} \B ChopperCommand-v0 & \B 1 & \B 333.4 & \B 559.6 & \B 0.02393 & \B 428.5 & \B 0.02393 \\
CrazyClimber-v0 & 0 & 124 & 195.7 & 0.1077 & 109.8 & 0.1077 \\
\rowcolor{gray!7} \B CrazyClimber-v0 & \B 1 & \B 506.3 & \B 716 & \B 0.1121 & \B 636.8 & \B 0.1121 \\
\B DemonAttack-v0 & \B 0 & \B 136.7 & \B 261.4 & \B 0.06215 & \B 198.9 & \B 0.06215 \\
\rowcolor{gray!7} DemonAttack-v0 & 1 & 111 & 195.7 & 0.06059 & 154.6 & 0.06059 \\
Enduro-v0 & 0 & 73.34 & 108.6 & 0.002759 & 96.77 & 0.002759 \\
\rowcolor{gray!7} \B Enduro-v0 & \B 1 & \B 88.7 & \B 149.7 & \B 0.003457 & \B 107.8 & \B 146.9 \\
FishingDerby-v0 & 0 & 12.47 & 22.56 & -5.809 & 18.55 & -0.1605 \\
\rowcolor{gray!7} \B FishingDerby-v0 & \B 1 & \B 20.02 & \B 26.59 & \B -0.8865 & \B 22.89 & \B -0.148 \\
Freeway-v0 & 0 & 0.0616 & 0.07526 & 0.005634 & 0.06213 & 0.005634 \\
\rowcolor{gray!7} \B Freeway-v0 & \B 1 & \B 4.473 & \B 5.675 & \B 0.009225 & \B 5.266 & \B 5.293 \\
\B Frostbite-v0 & \B 0 & \B 230.9 & \B 372.4 & \B 0.0794 & \B 216.8 & 0.0794 \\
\rowcolor{gray!7} Frostbite-v0 & 1 & 214.8 & 310.5 & 0.06663 & 209.4 & \B 299 \\
IceHockey-v0 & 0 & 0.5194 & 1.625 & -0.4062 & 0.5266 & 1.218 \\
\rowcolor{gray!7} \B IceHockey-v0 & \B 1 & \B 6.73 & \B 8.181 & \B -0.0109 & \B 7.421 & \B 7.618 \\
Jamesbond-v0 & 0 & 116.9 & 229.5 & \B 0.006727 & 166.9 & \B 0.006727 \\
\rowcolor{gray!7} \B Jamesbond-v0 & \B 1 & \B 129 & \B 246.5 & 0.006132 & \B 180.3 & 0.006132 \\
Kangaroo-v0 & 0 & 208.6 & 540.5 & \B 0.003812 & 39.76 & 536.6 \\
\rowcolor{gray!7} \B Kangaroo-v0 & \B 1 & \B 515.7 & \B 835.8 & 0.002641 & \B 700.7 & \B 835.4 \\
\B Krull-v0 & \B 0 & \B 1306 & \B 1741 & 0.3597 & \B 1435 & \B 1604 \\
\rowcolor{gray!7} Krull-v0 & 1 & 953.6 & 1332 & \B 0.4161 & 963.6 & 960.5 \\
KungFuMaster-v0 & 0 & 1.848 & 4.091 & \B 0.01861 & 1.737 & \B 0.01861 \\
\rowcolor{gray!7} \B KungFuMaster-v0 & \B 1 & \B 300 & \B 495.9 & 0.01443 & \B 338.4 & 0.01443 \\
Pitfall-v0 & 0 & -0.7866 & 0.002114 & -1.312 & -0.7986 & \B 0.002114 \\
\rowcolor{gray!7} \B Pitfall-v0 & \B 1 & \B 8.836 & \B 32.33 & \B -0.4345 & \B 9.447 & -0.01823 \\
\B Pong-v0 & \B 0 & \B 5.076 & \B 9.564 & -4.207 & \B 8.074 & \B -0.07481 \\
\rowcolor{gray!7} Pong-v0 & 1 & 4.766 & 7.895 & \B -1.264 & 5.027 & -0.07932 \\
Qbert-v0 & 0 & 341 & 492.7 & \B 0.0624 & 399.3 & \B 0.0624 \\
\rowcolor{gray!7} \B Qbert-v0 & \B 1 & \B 941.6 & \B 1377 & 0.06027 & \B 1072 & 0.06027 \\
\B Riverraid-v0 & \B 0 & \B 1115 & \B 1673 & \B 0.1119 & \B 1249 & 0.1119 \\
\rowcolor{gray!7} Riverraid-v0 & 1 & 938.8 & 1420 & 0.1079 & 1025 & \B 905 \\
Seaquest-v0 & 0 & 163.2 & 316.3 & \B 0.03345 & 147.4 & 0.03345 \\
\rowcolor{gray!7} \B Seaquest-v0 & \B 1 & \B 255.5 & \B 386.2 & 0.03208 & \B 323.4 & \B 331.2 \\
SpaceInvaders-v0 & 0 & 119.2 & 174.9 & \B 0.0773 & 150.7 & 165.8 \\
\rowcolor{gray!7} \B SpaceInvaders-v0 & \B 1 & \B 220.2 & \B 347.5 & 0.0721 & \B 282.1 & \B 284.9 \\
StarGunner-v0 & 0 & 2.04 & 3.595 & 0.0143 & 1.988 & 0.0143 \\
\rowcolor{gray!7} \B StarGunner-v0 & \B 1 & \B 5.32 & \B 30.72 & \B 0.01667 & \B 2.126 & \B 0.01667 \\
\B TimePilot-v0 & \B 0 & \B 52.56 & \B 272 & \B 0.03401 & 9.384 & 0.03401 \\
\rowcolor{gray!7} TimePilot-v0 & 1 & 47.64 & 82.9 & 0.03007 & \B 48.69 & \B 76.34 \\

\end{tabular}
\caption{Average Q-value reached during the training phase for each of the combinations of environment/masking. Best values are highlighted in bold.}
\label{table:results-q}
\end{table*}

\begin{table*}[h!]
\centering
\tiny
\begin{tabular}{ |l|l|l|l|l|l|l| } 

Environment & Mask & Mean & Maximum & Minimum & Median & Mod \\
Alien-v0 & 0 & 23.56 & 114 & \B 16 & 23 & 23 \\
\rowcolor{gray!7} \B Alien-v0 & \B 1 & \B 24.05 & \B 116 & \B 16 & \B 24 & \B 24 \\
\B Asterix-v0 & \B 0 & \B 48 & 279 & \B 33 & \B 48 & \B 49 \\
\rowcolor{gray!7} Asterix-v0 & 1 & 45.2 & \B 284 & 28 & 44 & 42 \\
\B Asteroids-v0 & \B 0 & \B 36.28 & 86 & \B 10 & \B 37 & \B 38 \\
\rowcolor{gray!7} Asteroids-v0 & 1 & 23.68 & \B 87 & 4 & 23 & 24 \\
Atlantis-v0 & 0 & 31.35 & \B 144 & \B 18 & 31 & 30 \\
\rowcolor{gray!7} \B Atlantis-v0 & \B 1 & \B 35.33 & 139 & 16 & \B 35 & \B 32 \\
\B BattleZone-v0 & \B 0 & \B 4347 & \B 5000 & \B 0 & \B 5000 & \B 5000 \\
\rowcolor{gray!7} BattleZone-v0 & 1 & 4182 & \B 5000 & \B 0 & \B 5000 & \B 5000 \\
BeamRider-v0 & 0 & 10.47 & 53 & \B 6 & \B 10 & \B 10 \\
\rowcolor{gray!7} \B BeamRider-v0 & \B 1 & \B 10.48 & \B 57 & \B 6 & \B 10 & \B 10 \\
Berzerk-v0 & 0 & 60.6 & 298 & 35 & 61 & 61 \\
\rowcolor{gray!7} \B Berzerk-v0 & \B 1 & \B 64.69 & \B 350 & \B 45 & \B 65 & \B 65 \\
Bowling-v0 & 0 & 2.183 & \B 11 & \B 1 & \B 2 & \B 2 \\
\rowcolor{gray!7} \B Bowling-v0 & \B 1 & \B 2.236 & \B 11 & \B 1 & \B 2 & \B 2 \\
Boxing-v0 & 0 & 2.877 & \B 14 & \B 2 & \B 3 & \B 3 \\
\rowcolor{gray!7} \B Boxing-v0 & \B 1 & \B 3.41 & \B 14 & \B 2 & \B 3 & \B 3 \\
\B Breakout-v0 & \B 0 & \B 68.69 & \B 676 & \B 38 & \B 64 & \B 59 \\
\rowcolor{gray!7} Breakout-v0 & 1 & 34.99 & 662 & 17 & 27 & 27 \\
ChopperCommand-v0 & 0 & 25 & \B 75 & 0 & 23 & 18 \\
\rowcolor{gray!7} \B ChopperCommand-v0 & \B 1 & \B 32.29 & 68 & \B 1 & \B 33 & \B 34 \\
CrazyClimber-v0 & 0 & 7.255 & 44 & \B 2 & 7 & 7 \\
\rowcolor{gray!7} \B CrazyClimber-v0 & \B 1 & \B 8.35 & \B 45 & \B 2 & \B 8 & \B 8 \\
\B DemonAttack-v0 & \B 0 & \B 22.41 & \B 121 & \B 2 & \B 22 & \B 24 \\
\rowcolor{gray!7} DemonAttack-v0 & 1 & 21.89 & 115 & \B 2 & \B 22 & \B 24 \\
Enduro-v0 & 0 & 1.482 & \B 7 & \B 0 & 1 & 1 \\
\rowcolor{gray!7} \B Enduro-v0 & \B 1 & \B 1.503 & \B 7 & \B 0 & \B 2 & \B 2 \\
FishingDerby-v0 & 0 & 2.654 & \B 13 & \B 2 & \B 3 & \B 3 \\
\rowcolor{gray!7} \B FishingDerby-v0 & \B 1 & \B 2.681 & 12 & \B 2 & \B 3 & \B 3 \\
\B Freeway-v0 & \B 0 & \B 2.447 & \B 12 & \B 2 & \B 2 & \B 2 \\
\rowcolor{gray!7} \B Freeway-v0 & \B 1 & \B 2.447 & \B 12 & \B 2 & \B 2 & \B 2 \\
Frostbite-v0 & 0 & 43.81 & 254 & 33 & 43 & 42 \\
\rowcolor{gray!7} \B Frostbite-v0 & \B 1 & \B 44.87 & \B 266 & \B 35 & \B 44 & \B 44 \\
\B IceHockey-v0 & \B 0 & \B 1.448 & \B 7 & \B 1 & \B 1 & \B 1 \\
\rowcolor{gray!7} IceHockey-v0 & 1 & 1.418 & \B 7 & \B 1 & \B 1 & \B 1 \\
Jamesbond-v0 & 0 & 40.69 & \B 309 & 21 & 38 & 31 \\
\rowcolor{gray!7} \B Jamesbond-v0 & \B 1 & \B 44.6 & 304 & \B 23 & \B 44 & \B 42 \\
\B Kangaroo-v0 & \B 0 & \B 28.49 & \B 161 & \B 20 & \B 27 & \B 25 \\
\rowcolor{gray!7} Kangaroo-v0 & 1 & 25.59 & 150 & 19 & 25 & 24 \\
Krull-v0 & 0 & 10.37 & 257 & \B 1 & \B 10 & \B 11 \\
\rowcolor{gray!7} \B Krull-v0 & \B 1 & \B 10.78 & \B 505 & 0 & \B 10 & 10 \\
\B KungFuMaster-v0 & \B 0 & \B 27.24 & 98 & \B 15 & \B 28 & \B 28 \\
\rowcolor{gray!7} KungFuMaster-v0 & 1 & 18.71 & \B 99 & 11 & 19 & 19 \\
\B Pitfall-v0 & \B 0 & \B 3478 & \B 5000 & \B 0 & \B 5000 & \B 5000 \\
\rowcolor{gray!7} Pitfall-v0 & 1 & 1827 & \B 5000 & \B 0 & 17 & \B 5000 \\
Pong-v0 & 0 & 2.919 & \B 26 & \B 1 & 2 & 2 \\
\rowcolor{gray!7} \B Pong-v0 & \B 1 & \B 3.134 & \B 26 & \B 1 & \B 3 & \B 3 \\
\B Qbert-v0 & \B 0 & \B 50.08 & 305 & \B 35 & \B 49 & \B 49 \\
\rowcolor{gray!7} Qbert-v0 & 1 & 47.74 & \B 312 & 31 & 47 & 47 \\
Riverraid-v0 & 0 & 34.71 & 137 & 21 & 33 & 32 \\
\rowcolor{gray!7} \B Riverraid-v0 & \B 1 & \B 38.37 & \B 138 & \B 24 & \B 37 & \B 36 \\
Seaquest-v0 & 0 & 27.13 & \B 193 & \B 17 & \B 27 & 26 \\
\rowcolor{gray!7} \B Seaquest-v0 & \B 1 & \B 27.75 & 183 & 16 & \B 27 & \B 27 \\
SpaceInvaders-v0 & 0 & 26.23 & 142 & 16 & 26 & 26 \\
\rowcolor{gray!7} \B SpaceInvaders-v0 & \B 1 & \B 29.88 & \B 147 & \B 17 & \B 30 & \B 31 \\
StarGunner-v0 & 0 & 24.98 & 115 & \B 12 & 25 & 24 \\
\rowcolor{gray!7} \B StarGunner-v0 & \B 1 & \B 26.26 & \B 119 & 10 & \B 26 & \B 26 \\
\B TimePilot-v0 & \B 0 & \B 19.24 & \B 73 & 6 & \B 18 & 16 \\
\rowcolor{gray!7} TimePilot-v0 & 1 & 17.83 & 68 & \B 9 & \B 18 & \B 17 \\

\end{tabular}
\caption{Average number of episodes played during the training phase for each of the combinations of environment/masking. Best values are highlighted in bold. Higher values are indicative of a exploratory strategy.}
\label{table:results-games}
\end{table*}

\clearpage

\vskip 0.2in

\bibliographystyle{elsarticle-harv}

\bibliography{bibliography}

\end{document}